\documentclass{article} 

\usepackage{arxiv}
\usepackage{amsfonts}
\usepackage{multirow}
\usepackage[numbers]{natbib}

\usepackage{amsmath,amsfonts,bm}

\def\eqref#1{(\ref{#1})}

\def\1{\bm{1}}

\DeclareMathAlphabet{\mathsfit}{\encodingdefault}{\sfdefault}{m}{sl}
\SetMathAlphabet{\mathsfit}{bold}{\encodingdefault}{\sfdefault}{bx}{n}

\newcommand{\E}{\mathbb{E}}

\DeclareMathOperator*{\argmax}{arg\,max}

\usepackage[title]{appendix}
\usepackage{hyperref}
\usepackage{url}
\usepackage{algorithm,algorithmic}
\usepackage{microtype}
\usepackage{graphicx}
\usepackage{natbib}
\usepackage{bbm}
\usepackage{booktabs}
\usepackage[english]{babel}
\usepackage[T1]{fontenc}
\usepackage{amsfonts}
\usepackage{subcaption}
\usepackage{graphicx}
\usepackage{adjustbox}
\usepackage[colorinlistoftodos]{todonotes}
\usepackage[utf8]{inputenc}
\usepackage{dsfont}
\usepackage{amsthm}
\usepackage{float}
\usepackage{amsfonts}
\usepackage{xcolor}
\usepackage{amsmath}
\usepackage{mathabx}
\usepackage{lipsum}

\newtheorem{proposition}{Proposition}[section]
\theoremstyle{definition}
\newtheorem{defn}{Definition}[section]

\newtheorem{exmp}{Example}[section]
\usepackage{xspace}

\usepackage{authblk}
\begin{document}
\title{On Completeness-aware Concept-Based \\ Explanations in Deep Neural Networks}
\date{}
\author[1]{Chih-Kuan Yeh}
\author[2]{Been Kim}
\author[3]{Sercan \"{O}. Ar{\i}k}
\author[3]{Chun-Liang Li}
\author[3]{Tomas Pfister}
\author[1]{Pradeep Ravikumar}
\affil[1]{Machine Learning Department, Carnegie Mellon University}
\affil[2]{Google Brain}
\affil[3]{Google Cloud AI}

\makeatletter
\def\BState{\State\hskip-\ALG@thistlm}
\makeatother
\newcommand{\Com}{\eta}
\newcommand{\Acc}{\textup{Acc}}
\newcommand{\con}{\mathbf{c}}
\newcommand{\conp}{\mathbf{p}}
\newcommand{\x}{\mathbf{x}}
\newcommand{\z}{\mathbf{z}}
\newcommand{\s}{\mathbf{s}}
\newcommand{\p}{f}
\newcommand{\h}{h}
\newcommand{\f}{\phi}
\newcommand{\var}{\textup{var}}
\newcommand{\cov}{\textup{cov}}
\newcommand{\cl}{\mathbf{\tau}}
\newcommand*{\medcup}{\mathbin{\scalebox{1.5}{\ensuremath{\cup}}}}%

\newlength\myindent
\setlength\myindent{2em}
\newcommand\bindent{%
  \begingroup
  \setlength{\itemindent}{\myindent}
  \addtolength{\algorithmicindent}{\myindent}
}
\newcommand\eindent{\endgroup}
\maketitle
\date{}

\begin{abstract}

Human explanations of high-level decisions are often expressed in terms of key concepts the decisions are based on. In this paper, we study such concept-based explainability for Deep Neural Networks (DNNs). First, we define the notion of \emph{completeness}, which quantifies how sufficient a particular set of concepts is in explaining a model's prediction behavior based on the assumption that complete concept scores are sufficient statistics of the model prediction. Next, we propose a concept discovery method that aims to infer a complete set of concepts that are additionally encouraged to be interpretable, which addresses the limitations of existing methods on concept explanations. To define an importance score for each discovered concept, we adapt game-theoretic notions to aggregate over sets and propose \emph{ConceptSHAP}. Via proposed metrics and user studies, on a synthetic dataset with apriori-known concept explanations, as well as on real-world image and language datasets, we validate the effectiveness of our method in finding concepts that are both complete in explaining the decisions and interpretable. \footnote{The code is released at https://github.com/chihkuanyeh/concept\_exp.}
\end{abstract}

\section{Introduction}
\vspace{-3mm}
The lack of explainability of deep neural networks (DNNs) arguably hampers their full potential for real-world impact. 
Explanations can help 
domain experts better understand rationales behind the model decisions,
identify systematic failure cases, and potentially provide feedback to model builders for improvements. 
Most commonly-used methods for DNNs explain each prediction by quantifying the importance of each input feature \citep{ribeiro2016should, lundberg2017unified}. One caveat with such explanations is that they typically focus on the local behavior for each data point, 
rather than globally explaining how the model reasons. Besides, the weighted input features 
are not necessarily the most intuitive 
explanations for human understanding, particularly when using low-level features such as raw pixel values. In contrast, human reasoning often comprise ``concept-based thinking,'' extracting similarities from numerous examples and grouping them systematically based on their resemblance~\citep{ARMSTRONG1983263, tenenbaum1999bayesian}. 
It is thus of interest to develop such ``concept-based explanations'' to characterize the global behavior of a DNN in a way understandable to humans, explaining how DNNs use concepts in arriving at particular decisions. 

A few recent studies have focused on bringing such concept-based explainability to DNNs, largely based on the common implicit assumption that the concepts lie in low-dimensional subspaces of some intermediate DNN activations. 
Via supervised training based on labeled concepts, TCAV \citep{kim2018interpretability} trains linear concept classifiers to derive concept vectors, and uses how sensitive predictions are to these vectors (via directional derivatives)
to measure the importance of a concept with respect to a specific class. 
\citet{zhou2018interpretable} considers the decomposition of model predictions in terms of projections onto concept vectors. 
Instead of human-labeled concept data, \citet{ghorbani2019towards} employs k-means clustering of super-pixel segmentations of images to discover concepts.
\citet{bouchacourt2019educe} proposes a Bayesian generative model involving concept vectors. 
One drawback of these approaches is that they do not take into account \textit{how much} each concept plays a role in the prediction. In particular, selecting a set of concepts salient to a particular class does not guarantee that these concepts are sufficient in explaining the prediction. The notion of sufficiency is also referred to as ``completeness'' of explanations, as in \citep{gilpin2018explaining, yang2019evaluating}. 
This motivates the following key questions: 
Is there an unsupervised approach to extract concepts that are 
sufficiently predictive of a DNN's decisions?
If so, how can we 
measure this sufficiency?

In this paper, we propose such a completeness score for concept-based explanations. 
Our metric can be applied to a set of concept vectors that lie in a subspace of some intermediate DNN activations, which is a general assumption in previous work in this context~\citep{kim2018interpretability,zhou2018interpretable}. 
Intuitively speaking, a set of ``complete'' concepts can fully explain the prediction of the underlying model. By further assuming that for a complete set of concepts, the projections of activations onto the concepts are a sufficient statistic for the prediction of the model, we may measure the ``completion'' of the concepts by the accuracy of the model just given these concept based sufficient statistics. For concept discovery, we propose a novel algorithm, which could also be viewed as optimizing a surrogate likelihood of the concept-based data generation process, motivated by topic modeling \citep{blei2003latent}. 
To ensure that the discovered complete concepts are also coherent (distinct from other concepts) and semantically meaningful, we further introduce an interpretability regularizer.

Beyond concept discovery, we also propose a score, \emph{ConceptSHAP}, for quantification of concept attributions as contextualized importance. ConceptSHAP uniquely satisfies a key set of axioms involving the contribution of each concept to the completeness score~\citep{shapley_1988, lundberg2017unified}. 
We also propose a class-specific version of ConceptSHAP that decomposes it with respect to each class in multi-class classification. This can be used to find class-specific concepts that contribute the most to a specific class. To verify the effectiveness of our \emph{automated} completeness-aware concept discovery method, we create a synthetic dataset with apriori-known ground truth concepts. We show that our approach outperforms all compared methods in correct retrieval of the concepts as well as in terms of its coherency via a user study.
Lastly, we demonstrate how our concept discovery algorithm provides additional insights into the behavior of DNN models on both image and language real-world datasets. 

\vspace{-3mm}

\section{Related Work}

\vspace{-3mm}
Most post-hoc interpretability methods fall under the categories: (a) feature-based explanation methods, that attribute the decision to important input features \citep{ribeiro2016should, lundberg2017unified,smilkov2017smoothgrad, l_c_shapley}, (b) sample-based explanation methods, that attribute the decision to previously observed samples~\citep{koh2017understanding,yeh2018representer,khanna2019interpreting, attention_prototypical}, and (c) counterfactual-based explanation methods, which answers the question: ``what to alter in the current input to change the outcome of the model''~\citep{Wachter2017Counterfactual,dhurandhar2018explanations,Hendricks2018GroundingVE,vanderwaa2018,goyal2019counterfactual,Joshi2019Towards, Poyiadzi2020FACE}.Recent work has also focused on \emph{evaluations} of explanations, ranging from human-centric evaluations \citep{lundberg2017unified, kim2018interpretability, lage2019human} to functionally-grounded evaluations \citep{samek2016evaluating,kim2016examples,ancona2017towards,yeh2019on, yang2019evaluating, guidotti2018survey, hooker2018evaluating, yang2019benchmarking}. Our work provides an evaluation of concept explanations based on the completeness criteria, which is related to the `fidelity' \citep{guidotti2018survey}.


Our work is related to methods that learn semantically-meaningful latent variables. Some use dimensionality reduction methods~\citep{chan2015pcanet, kingma2013auto}, while others uncover higher level human-relatable concepts by dimensionality reduction (e.g. for speech~\citep{chorowski2019unsupervised} and language~\citep{unsupervised_sentiment, 8668423}). More recently~\citet{locatello2018challenging} shows that meaningful latent dimensions cannot be acquired in a completely unsupervised setting, implying the necessity of inductive biases for discovering meaningful latent dimensions. Our work uses indirect supervision from the classifier of interest to discover semantically meaningful latent dimensions. \citet{chen2019looks} uses the representative training patches to explain a prediction in a self-interpretable framework for image classification. \citet{koh2020concept} learn models that first predict human labeled concepts, then use concept scores to predict model. \citet{goyal2019explaining} measures the causal effect of concepts by using a conditional VAE model. These methods either require a training model from scratch or training a generative model, whereas our method can be applied on given models and different data types.
\vspace{-2mm}
\section{Defining Completeness of Concepts}
\vspace{-2mm}
\paragraph{Problem setting:}
Consider a set of $n$ training examples $\x^1, \x^2, ..., \x^n$, corresponding labels $y^1, y^2, ..., y^n $ and a given pre-trained DNN model that predicts the corresponding $y$ from the input $\x$. We assume that the pre-trained DNN model can be decomposed into two functions: the first part $\Phi(\cdot)$ maps input $\x^i$ into an intermediate layer $\Phi(\x^i)$, and the second part $h(\cdot)$ maps the intermediate layer $\Phi(\x^i)$ to the output $h(\Phi(\x^i))$, which is a probability vector for each class, and $h_y(\Phi(\x))$ is the probability of data $\x$ being predicted as label $y$ by the model $\p$. For DNNs that build up by processing parts of input at a time, such as those composed of convolutional layers, we can additionally assume that $\Phi(\x^i)$ is the concatenation of $[\phi(\x^i_1), ...,  \phi(\x^i_T)]$, such that $\Phi(\cdot) \in \mathbb{R}^{(T \cdot d)}$, and $\phi(\cdot) \in \mathbb{R}^{d}$. Here, $\x^i_1, \x^i_2, ..., \x^i_T$ denote different, potentially overlapping parts of the input for $\x^i$, such as a segment of an image or a sub-sentence of a text. These parts for example, can be chosen to correspond to the receptive field of the neurons at the intermediate layer $\Phi(\cdot)$. We will use these $\x^i_t$ to relate discovered concepts. 
As an illustration of such parts, consider the fifth convolution layer of a VGG-16 network with input shape $224\times 224$ have the size $7\times 7 \times 512$. If we treat this layer as $\Phi(\x^i)$, $\phi(\x^i_1)$ corresponds to the first 512 dimensions of the intermediate layer (with size $7\times 7 \times 512$), and $\Phi(\x^i) = [\phi(\x^i_1), ...,  \phi(\x^i_{49})]$. Here, each $\x^i_j$ corresponds to a $164 \times 164$ square in the input image (with effective stride 16), which is the receptive field of convolution layer 5 of VGG-16 \citep{araujo2019computing}.
We note that when the receptive field of $\phi(\cdot)$ is equal to the entire input size, such as for multi-layer perceptrons, we may simply choose $T=1$ so that $\x^i_{1:T} = \x^i$ and $\Phi(\x^i) = \phi(\x^i_1)$. Thus, our method can also be generally applied to any DNN with an arbitrary structure besides convolutional layers. To choose the intermediate layer to apply concepts, we follow previous works on concept explanations \cite{kim2018interpretability, ghorbani2019towards} by starting from the layer closest to the prediction until we reached a layer that user is happy with, as higher layers encodes more abstract concepts with larger receptive field, and lower layers encodes more specific concepts with smaller receptive field.




Suppose that there is a set of $m$ concepts denoted by unit vectors\footnote{We apply additional normalization to $\phi(\cdot)$ so it has unit norm and keep the notation for simplicity.} $\con_1, \con_2, ... , \con_m $ that represent linear directions in the activation space $\f(\cdot) \in \mathbb{R}^d$, given by a concept discovery algorithm. 
For each part of data point $\x_t$ (We omit $i$ for notational simplicity), 
The inner product between the data and concept vector is viewed as the closeness of the input $\x_t$ and the concept $\con$ following \cite{kim2018interpretability, ghorbani2019towards}. If $\langle\phi(\x_t),c_j\rangle$ is large, then we know that $\x_t$ is close to concept $j$. However, when $\langle\phi(\x_t),c_j\rangle$ is less than some threshold, the dot product value is not semantically meaningful other than the the input is not close to the concept. Based on this motivation, we define the \emph{concept product} for part of data $\x_t$ as $v_\con(\x_t) := \text{TH}(\langle\phi(\x_t),c_j\rangle, \beta)_{j=1}^{m} \in \mathbb{R}^m$, where $\text{TH}$ is a threshold which trims value less than $\beta$ to 0. We normalize the concept product to unit norm for numerical stability, and aggregate upon all parts of data to obtain the \emph{concept score} for input $\x$ as $ v_\con(\x) = (\frac{v_c(x_t)}{\|v_c(x_t)\|_2})_{t=1}^{T} \in \mathbb{R}^{T \cdot m}$. 

We assume that for ``sufficient'' concepts, the concept scores should be sufficient statistics for the model output, and thus we may evaluate the completeness of concepts by how well we can recover the prediction given the concept score. Let $g: \mathbb{R}^{T\cdot m} \rightarrow \mathbb{R}^{T \cdot d}$ denote any mapping from the concept score to the activation space of $\Phi(\cdot)$. If concept scores $v_\con(\cdot)$ are sufficient statistics for the model output, then there exists some mapping $g_\p$ such that $h(g_\p(v_\con(\x))) \approx \p(\x)$. We can now formally define the completeness core for a set of concept vectors $\con_1, ..., \con_m$:

\begin{defn}\label{df:com}
\textbf{Completeness Score:} Given a prediction model $\p(\x) = \h( \f(\x))$, a set of concept vectors $\con_1, ..., \con_m$, we define the completeness score $\Com_\p(\con_1,..., \con_m)$ as:
	    \begin{equation} \label{eq:complete}
         \Com_\p(\con_1,..., \con_m) = \frac{\sup_g \mathbb{P}_{\x,y \sim V}[{y} = \argmax_{y'} h_{y'}(g(v_\con(\x)))]-a_r}{\mathbb{P}_{\x,y \sim V}[{y} = \argmax_{y'}\p_{y'}(\x)]-a_r} ,
	    \end{equation}
where $V$ is the set of validation data and $\sup_g \mathbb{P}_{\x,y \sim V}[{y} = \argmax_{y'} h_{y'}(g(v_\con(\x)))]$ is the best accuracy by predicting the label just given the concept scores $v_\con(\x)$, and $a_r$ is the accuracy of random prediction to equate the lower bound of completeness score to 0. When the target $y$ is multi-label, we may generalize the definition of completeness score by replacing the accuracy with the binary accuracy, which is the accuracy where each label is treated as a binary classification.

\end{defn}

To calculate the completeness score, we can set $g$ to be a DNN or a simple linear projection, and optimize using stochastic gradient descent. In our experiments, we simply set $g$ to be a two-layer perceptron with 500 hidden units. We note that we approximate $\p(\x_t)$ by $h(g_\p(v_\con(\x_t)))$, but not an arbitrary neural network $h_g(v_\con(\x_t))$ for two benefits: (a) the measure of completeness considers the architecture and parameter of the given model to be explained (b) the computation is much more efficient since we only need to optimize the parameters of $g$, instead of the whole backbone $h_g$.
The completeness score measures how ``sufficient'' are the concept scores as a sufficient statistic of the model, based on the assumption that the concept scores of ``complete'' concepts are sufficient statistics of the model prediction $\p(\cdot)$. By measuring the accuracy achieved by the concept score, we are effectively measuring how ``complete'' the concepts are. We note that the completeness score can also be used to measure how sufficient concepts can explain a dataset independent of the model, by replacing $\phi(\cdot), h(\cdot)$ with identical functions, and $\p(\x)$ with $y$. Below is an illustrative example on why we need the completeness score:
\begin{exmp} \label{ex:com}
Consider a simplified scenario where we have the input $\x \in \mathbb{R}^m$, and the intermediate layer $\Phi$ is the identity function. In this case, the $m$ concepts $\con_1, \con_2, ..., \con_m$ are the one-hot encoding of each feature in $\x$. Assume that the concepts $\con_1, \con_2, ..., \con_m$ follow independent Bernoulli distribution with $p = 0.5$, and the model we attempt to explain is $f(\x) = \con_1 \text{ XOR } \con_2 ... \text{ XOR } \con_m$. The ground truth concepts that are sufficient to the model prediction should then be $\con_1, \con_2, ..., \con_m$. However, if we have the information on $\con_1, \con_2, ..., \con_{m-1}$ but do not have information on $\con_m$, we may have at most $0.5$ probability to predict the output of the model, which is the same as the accuracy of random guess. In this case, $\Com_\p(\con_1, \con_2, ..., \con_{m-1})=0$. On the other hand, given $\con_1, \con_2, ..., \con_m$, $\Com_\p(\con_1, \con_2, ..., \con_{m})=1$.
\end{exmp}

The completeness score offers a way to assess the `sufficiency' of the discovered concepts to ``explain" reasoning behind a model's decision. 
Not only the completeness score is useful in evaluating a proposed concept discovery method, but it can also shed light on how much of the learned information by DNN may not be `understandable' to humans. For example, if the completeness score is very high, but discovered concepts aren't making cohesive sense to humans, this may mean that the DNN is basing its decisions on other concepts that are potentially hard to explain. 

\vspace{-2mm}
\section{Discovering Completeness-aware Interpretable Concepts}
\vspace{-2mm}
\subsection{Limitations of existing methods}
\vspace{-2mm}
Our goal is to discover a set of maximally-complete concepts under the definition \ref{df:com}, where each concept is interpretable and semantically-meaningful to humans. We first discuss the limitations of recent notable works related to concept discovery and then explain how we address them. 
TCAV and ACE are concept discovery methods that use training data for specific concepts and use trained linear concept classifier to derive concept vectors. They quantify the saliency of a concept to a class using `TCAV score', based on the similarity of the loss gradients to the concept vectors. This score implicitly assumes a first-order relationship between the concepts and the model outputs. Regarding labeling of the concepts, TCAV relies on human-defined labels, while ACE uses automatically-derived image clusters by k-means clustering of super-pixel segmentations. There are two main caveats to these approaches. The first is that while they may retrieve an important set of concepts, there is no guarantee on how `complete' the concepts are in explain the model -- e.g., one may have 10 concepts with high TCAV scores, but they may still be very insufficient in understanding the predictions. Besides, human-suggested exogenous concept data might even encode confirmation bias. The second caveat is that their saliency scores may fail to capture concepts that have non-linear relationships with the output due to first-order assumption. The concepts in Example \ref{ex:com} might not be retrieved by the TCAV score since $\text{XOR}$ is not a linear relationship. Overall, our completeness score complements previous works in concept discovery by adding a criterion to determine whether a set of concepts are sufficient to explain the model. The discussion of our relation to PCA is in the Appendix.

\vspace{-2mm}
\subsection{Our method}
\vspace{-2mm}
The goal of our method is to obtain concepts that are \emph{complete} to the model. We consider the case where each data point $\x^i$ has parts $\x_{1:T}^i$, as described above. We assume that input data has spatial dependency, which can help learning coherent concepts. Thus, we encourage proximity between each concept and its nearest neighbors patches. Note that the assumption works well with images and language, as we will demonstrate in the result section. We aim that the concepts would obtain consistent nearest neighbors that only occur in parts of the input, e.g. head of animals or the grass in the background so that the concepts are pertained to certain spacial regions. By encouraging the closeness between each concept and its nearest neighbors, we aim to obtain consistent nearest neighbors to enhance interpretability. Lastly, we optimize the completeness terms to encourage the \emph{completeness} of the discovered concepts.

\vspace{-2mm}
\paragraph{Learning concepts:}
\label{sec:phi}
To optimize the completeness of the discovered concepts, we optimize the surrogate loss for the completeness term for both concept vectors $\con_{1:m}$ and the mapping function $g$:
\begin{equation} \label{eq:opt_con}
    \argmax_{\con_{1:m},g} \log \mathbb{P}[ h_{y}(g(v_\con(\x)))]
\end{equation}

An interpretation for finding the underlying concepts whose concept score maximizes the recovered prediction score is analogous to treating the prediction of DNNs as a topic model. By assuming the data generation process of $(\x,y)$ follows the probabilistic graphical model $\x_t \rightarrow \z_t$ and $\z_{1:T} \rightarrow y$, such that the concept assignment $\z_t$ is generated by the data, and the overall concept assignment $\z_{1:T}$ determines the label $y$. The log likelihood of the data $\log P[y|\x]$ can be estimated by $\log P[y|\x] = \log \int_z P[y|z] P[z|\x] \approx  \log P[y|E[z|\x]],$ by replacing the sampling by a deterministic average. We note that $v_\con(\x_{1:T})$ resembles $E[z|\x]$ and $P(y|h(g(v_\con(\x)))$ resembles $P[y|E[z|\x]]$, and as in supervised topic modeling \cite{mcauliffe2008supervised}, we jointly optimize the latent ``topic'' and the prediction model, but in an end-to-end fashion to maintain efficiency instead of EM update.


To enhance the interpretability of our concepts beyond ``topics'', we further design a regularizer to encourage the spacial dependency (and thus coherency) of concepts.
Intuitively, we require that the top-K nearest neighbor training input patches of each concept to be sufficiently close to the concept, and different concepts are as different as possible. 
This formulation encourages the top-K nearest neighbors of the concepts would be coherent, and thus allows explainability by ostensive definition. K is a hyperparameter that is usually chosen based on domain knowledge of the desired frequency of concepts. In our results, we fix K to be half of the average class size in our experiments. When using batch update, we find that picking $K = (\text{batch size} \cdot \text{average class ratio})/2$ works well in our experiments, where $\text{average class ratio} = {\text{average instance of each class}}/{ \text{total number of instances}}$. That is, the regularizer term tries to maximize $\Phi(\x_t^i)\cdot \con_k$ while minimizing $\con_j\cdot \con_k$. $\Phi(\x_t^i)\cdot \con_k$ is the similarity between the $t^{th}$ patch of the $i^{th}$ example and $\con_j\cdot \con_k$ is the similarity between the $j^{th}$ concept vector and the $k^{th}$ concept vector. By averaging over all concepts, and defining $T_{\con_k}$ as the set of top-K nearest neighbors of $\con_k$, the final regularization term is \[ R(\con) = \lambda_1 \frac{\sum_{k=1}^m  \sum_{\x_a^b \in T_{\con_k}}\Phi(\x_a^b)\cdot \con_k}{mK}   - \lambda_2  \frac{\sum_{j \neq k}\con_j\cdot \con_k}{{m(m-1)}}.\]

By adding the regularization term to \eqref{eq:opt_con}, the final objective becomes
\begin{equation}\label{eq:opt_final}
    \argmax_{\con_{1:m},g}\log P(h_y(g(v_\con(\x_{1:T}))) +  R(\con),
\end{equation}

for which we use stochastic gradient descent to optimize variables $\con_{1:m}, g$ jointly. When the optimization converges, $g$ is a (local) optimal value given $\con_{1:m}$. Since only concept vectors $\con_{1:m}$, and the mapping function $g$ is optimized in the process, the optimization process converges much faster compared to training the model from scratch. The computational cost for discovering concepts and calculating conceptSHAP is about 3 hours for AwA dataset and less than 20 minutes for the toy dataset and IMDB, using a single 1080 Ti GPU, which can be further accelerated with parallelism. The choice of which layer to apply $h_y, g$ and the corresponding architecture are further discussed in the appendix.

\vspace{-2mm}
\subsection{ConceptSHAP: How important is each concept?}
\vspace{-2mm}
Given a set of concept vectors $C_S = \{\con_1, \con_2, ... \con_{m}\}$ with a high completeness score, we would like to evaluate the importance of each individual concept by quantifying how much each individual concept contributes to the final completeness score. Let $\s_i$ denote the importance score for concept $\con_i$, such that $\s_i$ quantifies how much of the completeness score $\Com(C_S)$ is contributed by $\con_i$. Motivated by its successful applications in quantifying attributes for complex systems, we adapt Shapley values \citep{shapley_1988} to fairly assign the importance of each concept (which we call ConceptSHAP):
\begin{defn}\label{df:comshap}
    Given a set of concepts $C_S = \{\con_1, \con_2, ... \con_{m}\}$ and some completeness score $\Com$, we define the ConceptSHAP $\s_i$ for concept $\con_i$ as 
\begin{equation*}
\begin{split}
     \s_i(\Com) &= \sum\nolimits_{S \subseteq C_s \text{\textbackslash} \con_i} \frac{(m-|S|-1)!|S|!}{m!}[\Com(S \medcup \{\con_i\}) - \Com(S)],
\end{split}
\end{equation*}
\vspace{-5mm}
\end{defn}
The main benefit of Shapley for importance scoring is that it uniquely satisfies the set of desired axioms: efficiency, symmetry, dummy, and additivity. As these axioms are widely discussed in previous works \citep{shapley_1988, lundberg2017unified}, we leave the definitions and proof to Appendix.
\vspace{-3mm}
\paragraph{Per-class saliency of concepts:}
Thus far, conceptSHAP measures the global attribution (i.e., contribution to completeness when all classes are considered). However, per-class saliency, how much concepts contribute to prediction of a particular class, might be informative in many cases. To obtain the concept importance score for each class, we define the completeness score with respect to the class by considering data points that belong to it, which is formalized as:
\begin{defn}\label{df:comshapclass}
	Given a prediction model $\p(\x) = \h( \f(\x))$, a set of concept vectors $\con_1, \con_2, ..., \con_m$ that lie in the feature subspace in $\f(\cdot)$, we define the completeness score $ \Com_{\p,j}(\con_1,..., \con_m)$ for class $j$ as:
	    \begin{equation} 
        \Com_{\p,j}(\con_1,..., \con_m) = \frac{\mathbb{P}_{\x,y \in V_j}[{y} = \argmax_{y'} h_{y'}(\hat{g}(v_\con(\x)))]-a_{r,j}}{\mathbb{P}_{\x,y \in V}[y = \argmax_{y'}\p_{y'}(\x_{1:T})]-a_r} ,
        \end{equation}
\end{defn}
where $V_j$ is the set of validation data with ground truth label $j$, and $a_{r,j}$ is the accuracy of random predictions for data in class $j$, and $\hat{g}$ is derived via the optimization of completeness. We then define the perclass ConceptSHAP for concept $i$ with respect to class $j$ as:
\begin{defn}\label{df:conceptshap}
	Given a prediction model $\p(\x)$, a set of concept vectors in the feature subspace in $\f(\cdot)$. We can define the perclass ConceptSHAP for concept $i$ with respect to class $j$ as:
        $\s_{i,j}(\Com) = \s_i(\Com_{\p,j}).$
\end{defn}
\vspace{-1.2mm}
For each class $j$, we may select the concepts with the highest conceptSHAP score with respect to class $j$. We note that $\sum_j \frac{|V_j|}{|V|}\Com_{\p,j} = \Com$ and thus with the additivity axiom, $\sum_j\frac{|V_j|}{|V|} \s_{i,j}(\Com_{\p,j}) = \s_i(\Com)$. 
\vspace{-2mm}
\section{Experiments}
\vspace{-2mm}

In this section, we demonstrate our method both on a synthetic dataset, where we have ground truth concept importance, as well as on real-world image and language datasets. 
\begin{figure*}
\vspace{-2mm}
\centering
\begin{subfigure}{.48\textwidth}
  \centering
  \includegraphics[width=0.88\linewidth]{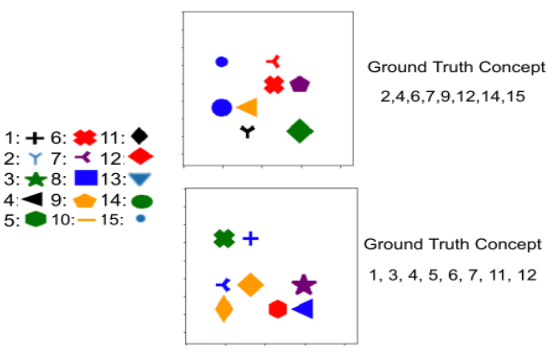}
  \vspace{-1mm}
  \caption{Two random images and corresponding ground truth concepts (with their legend on the left) -- each object corresponds to a ground truth concept solely via the shape information.}
  \label{fig:toy_dem}
 \end{subfigure}
 \hspace{3mm}
\begin{subfigure}{.47\textwidth}
  \centering
  \includegraphics[width=1.0\linewidth]{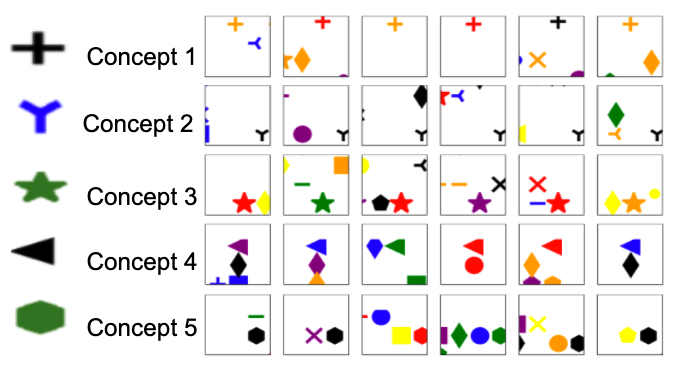}
  \vspace{-2.5mm}
  \caption{Top nearest neighbors (each neighbor corresponds to a part of the full image) of each discovered concepts. The ground truth concepts, determined by their shape (with random colors), are on the left.}
  \label{fig:toy_nn_new}
\end{subfigure}

\caption{ Examples (left) and nearest neighbors of our method (right) on Synthetic data.}
\vspace{-2mm}
\label{fig:toy_concept}
\end{figure*}

\vspace{-2mm}
\subsection{Synthetic data with ground truth concepts}
\vspace{-2mm}
\paragraph{Setting:} We construct a synthetic image dataset with known and complete concepts, to evaluate how accurately the proposed concept discovery algorithm can extract them. In this dataset, each image contains at most 15 shapes (shown in Fig.~\ref{fig:toy_dem}), and only 5 of them are relevant for the ground truth class, by construction.
For each sample $\x^i$, $\z_{j}^i$ is a binary variable which represents whether $\x^i$ contains shape $j$. $\z_{1:15}^i$ is a 15-dimensional binary variable with elements independently sampled from Bernoulli distribution with $p=0.5$.
We construct a 15-dimensional multi-label target for each sample, where the target of sample $i$, $y^i$ is a function that depends only on $\z_{1:5}^i$, which represents whether the first 5 shape exists in $\x^i$. For example, $y_1 = \sim( \mathbf{\z_1} \cdot \mathbf{\z_3})+ \mathbf{\z_4}, y_2 =  \mathbf{\z_2}+ \mathbf{\z_3}+ \mathbf{\z_4}, y_3 =  \mathbf{\z_2}\cdot \mathbf{\z_3} + \mathbf{\z_4}\cdot \mathbf{\z_5}$, where $\sim$ denotes logical Not (details are
in Appendix). 
We construct 48k training samples and 12k evaluation samples and use a convolutional neural network with 5 layers, obtaining $0.999$ accuracy. We take the last convolution layer as the feature layer $\f(\x).$ 




\begin{figure*}[b]
\vspace{-1.5mm}
\centering
\begin{subfigure}{.48\textwidth}
  \centering
   \includegraphics[width=1.02\linewidth]{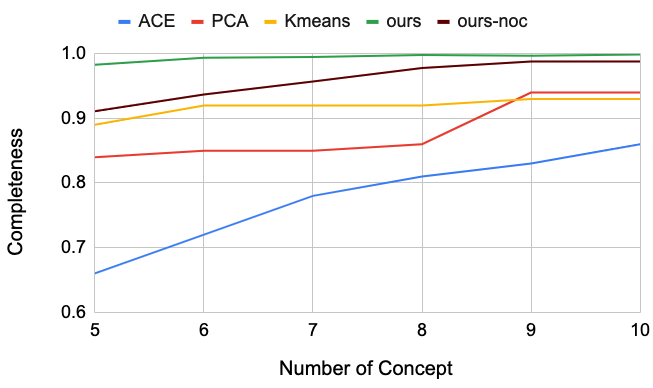}
  \label{fig:sub2}
\end{subfigure}
\begin{subfigure}{.48\textwidth}
  \centering
   \includegraphics[width=1.02\linewidth]{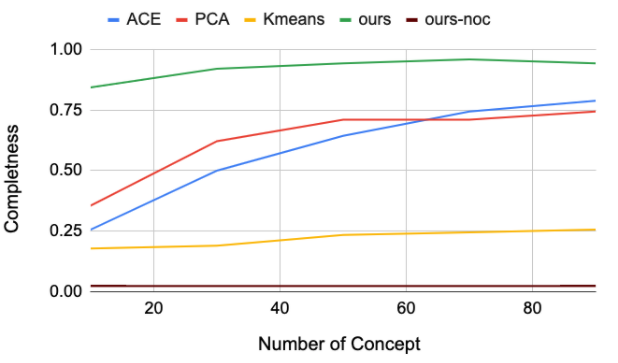}
  \label{fig:sub3}
\end{subfigure}
\vspace{-6mm}
\caption{Completeness scores on synthetic dataset (left) and completeness scores on AwA (right) versus different number of discovered concepts $m$ for all concept discovery methods in the synthetic dataset.
Ours-noc refers to our method without the completeness score objective as an ablation study.}
\vspace{-5mm}
\label{fig:completeness}
\end{figure*}
\vspace{-2mm}

\vspace{-2mm}
\paragraph{Evaluations:} 
We conduct a user-study with 20 users to evaluate the nearest neighbor samples of a few concept discovery methods. At each question, a user sees 10 nearest neighbor images of each discovered concept vector (as shown on the right of Fig. \ref{fig:toy_nn_new}), and is asked to choose the most common and coherent shape out of the 15 shapes based on the 10 nearest neighbors. We evaluate the results for our method, k-means clustering, PCA, ACE, and ACE-SP when $m=5$ concepts are retrieved. Each user is tested on two randomly chosen methods in random order, and thus each method is tested on 8 users. We report the average number of correct concepts and the number of agreed concepts (where the mode of each question is chosen as the correct answer) for each method answered by users in Table \ref{table:human}. The average number of correct concepts measures how many of the correct concepts are retrieved by user via nearest neighbors. The average number of agreed concepts measures how consistent are the shapes retrieved by different users, which is related to the coherency and conciseness of the nearest neighbors for each method. We also provide an automated alignment score based on how the discovered concept direction classifies different concepts -- see Appendix for details.


\vspace{-3mm}
\paragraph{Results:}
We compare our methods to ACE, k-means clustering, and PCA. For k-means and PCA, we take the embedding of the patch as input to be consistent to our method. For ACE, we implement a version which replaces the superpixels with patches and another version that takes superpixels as input, which we refer as ACE and ACE-SP respectively. We report the correct concepts and agreed concepts from the user study, and an automated alignment score which does not require humans. We do not calculate the alignment score of ACE-SP since it does not operate on patches and thus is unfair to compare with others (which would lead to much lower scores.) Our method outperforms others on corrected concepts and alignment score, is superior in retrieving the accurate concepts beyond the limitations of others. The number of agreed concepts is also the highest for our method, showing how highly-interpretability it is to humans such that the same concepts are consistently retrieved based on nearest neighbors.
As qualitative results, Fig. \ref{fig:toy_nn_new} shows the top-6 nearest neighbors for each concept $\con_k$ of our concept discovery method based on the dot product $\langle\con_k , \Phi(\x_a)^b\rangle$. All nearest neighbors contain a specific shape that corresponds to the ground-truth shapes 1 to 5. For example, all nearest neighbors of concept 1 contain the ground truth shape 1, which are cross as listed in Fig. \ref{fig:toy_dem}. A complete list of the top-10 nearest neighbors of all concept discovery methods is shown in Appendix. 
\begin{table}[!t]
\caption{The average number of correct and agreed concepts by users based on nearest neighbors.}
\centering
\adjustbox{max width=0.9\linewidth}{
\centering
\begin{tabular}{@{}lcccccc@{}}
\toprule
&  ACE & ACE-SP & PCA & k-means & \textbf{Ours} \\ \midrule
correct concepts&   $3.0 \pm 0$  &  $2.75\pm0.46$  &   $3.875\pm0.35$   & $3.75\pm0.46$   & \bm{$5.0 \pm 0$}     \\
agreed concepts&   $4.625$  &  $4.75$ &  $4.375$    & $4.75$    & \bm{$5.0$}  \\
automated alignment &   $0.741$  &  $-$ &  $0.876$    & $0.864$    & \bm{$0.98$} 
\\
 \bottomrule
\end{tabular}
}
\label{table:human}
\vspace{-6mm}
\end{table}
\begin{figure*}[b]
\vspace{-5mm}
\centering
  \includegraphics[width=0.99\linewidth]{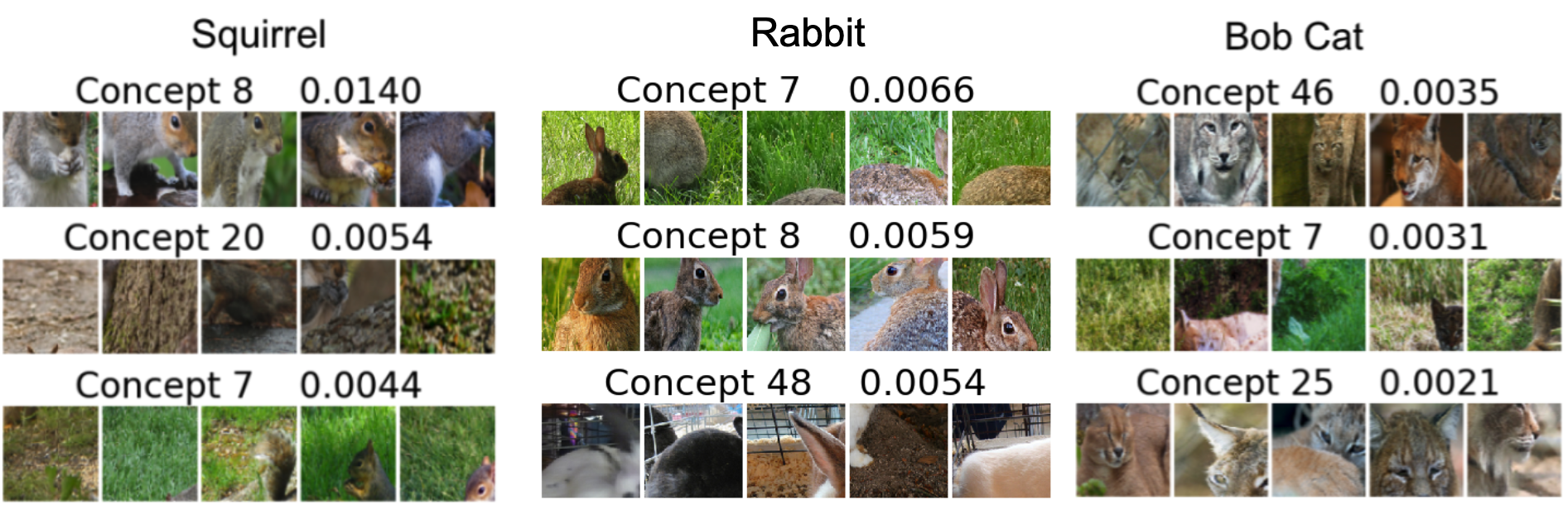}
  \vspace{-2mm}
  \caption{Concept examples with the samples that are the nearest to concept vectors in the activation space in AwA. The per-class ConceptSHAP score is listed above the images.
  }
  \vspace{-4mm}
  \label{fig:awa_new}
  
\end{figure*}

\vspace{-3mm}
\subsection{Image classification}
\vspace{-3mm}
\paragraph{Setting and metrics:}
We perform experiments on Animals with Attribute (AwA) \citep{lampert2009learning} that contains 50 animal classes. We use 26905 images for training and 2965 images for evaluation. 
We use the Inception-V3 model, pre-trained on Imagenet \citep{szegedy2016rethinking}, which yields $0.9$ test accuracy. We apply our concept discovery algorithm to obtain $m=70$ concepts. We conduct ad-hoc duplicate concept removal, by removing one concept vector if there are two vectors where the dot product is over 0.95. 
This gives us 53 concepts in total. 
We then calculate the ConceptSHAP and per class saliency score for each concept and each class. For each class, the top concepts based on the conceptSHAP are the most important concepts to classify this class, as shown in Fig.\ref{fig:awa_new}. While ConceptSHAP is useful in capturing the sufficiency of concepts for prediction, sometimes we may 
want to show examples. We propose to measure the quality of the nearest neighbors explanations by the average dot product between the nearest-neighbor patches that belongs to the class and the concept vector. In other words, the quality of the nearest neighbors explanations is simply the first term in $R(\con)$, which we denote as $R_1(\con) = \sum_{k=1}^m \sum_{\x_a^b \subseteq  T_{\con_k}}\langle\Phi(\x_a^b), \con_k\rangle$, where the top-K set is limited to image patches in the class of interest. When the nearest neighbor set contains patches of the same original image, we only show the patch with the highest similarity to the concept to increase the diversity. 

\vspace{-3mm}

\paragraph{Results:} We show the top concepts (ranked by conceptSHAP) of 3 classes with $R_1(\con)$>0.8 in Fig. \ref{fig:awa_new} (full results are in Appendix).
Note that since our method finds concepts for \textit{all} classes as opposed to specific to one class (such as \cite{ghorbani2019towards, chen2019looks}), we discover common concepts across many classes. For example, concept 7, whose nearest neighbors show grass texture, is important for the classes `Squirrel', `Rabbit', `Bob Cat', since all these animals appear in prairie. 
Concept 8 shows a oval head and large black round eyes shared by the classes `Rabbit', `Squirrel', `Weasel', while concept 46 shows head of the `Bob cat', which is shared by the classes `Lion', `Leopard', and `Tiger', `Antelope', and `Gorilla', which all show animal heads that are more rectangular and significantly different to the animal heads of concept 8. We find that having concepts shared between classes is useful to interpret the model.
Fig.~\ref{fig:completeness} shows that our method achieves the highest completeness of all methods on both the synthetic dataset and AwA. As a sanity check, we include the baseline `ours-noc', where the completeness objective is removed from \eqref{eq:opt_final}. Our method has much higher completeness than `ours-noc', demonstrating the necessity of the completeness term. In Appendix, we show some more top concepts for PCA, Kmeans, where the top concepts for PCA and Kmeans are also chosen as the same setting as ours.

\paragraph{Human Study:} We conduct a human study for the top neighbors for concepts discovered by our method, PCA, and Kmeans on the classes `Squirrel', `Rabbit', `Bob Cat'. For each method, we randomly choose 1 top concept per class for the 3 classes, and thus we choose 3 concepts per method (9 concepts in total). For each concept, we show users 4 top images of that concept, and ask users to choose the image (out of 3 different options) that they believe should belong to the same concept (where one of the option will actually belong to the same concept, and the other two are random image patches of the same class that does not belong to that concept). We then calculate the average accuracy to measure the human interpretability of the concept discover method. We conduct a user study with 10 users, where each of them are asked with the same 9 questions (1 question per concept chosen). The average correct ratio for our method, PCA, and Kmeans are 0.733, 0.267, and 0.6 respectively, showing our method's superiority. Kmeans outperforms PCA as it also encourages closeness of top nearest neighbors (which is better for ostensive definition).

\vspace{-2mm}
\subsection{Text classification}
\vspace{-2mm}
\begin{table}[!t]
\caption{The 4 discovered concepts and some nearest neighbors along with the most frequent words that appear in top-500 nearest neighbors.}
\centering
\adjustbox{max width=1.0\linewidth}{
\centering
\begin{tabular}{@{}lccc@{}}
\toprule
  Concept & Nearest Neighbors & Frequent words & ConceptSHAP \\ \midrule
   & poorly constructed what comes across as interesting is the& worst (168) ever (69) movie (61) seen (55)    & \\
   1 & wasting my time with a comment but this movie &  film (50) awful (42) time(40) waste (34) & 0.280\\
   & awful in my opinion there were <UNK> and the &  poorly (26) movies (24) films (18) long (17) & \\\midrule
    & normally it would earn at least 2 or 3 & not (58) movie (39) make (25) too (23) & \\
   2& <UNK> <UNK> is just too dumb to be called & film (22) even (19) like (18) 2 (16)  & 0.306 \\
   & i feel like i was ripped off and hollywood & never (14) minutes (13) 1 (12) doesn't (11) & \\\midrule
   & remember awaiting return of the jedi with almost <UNK> & movies (19) like (18) see (16) movie (15) & \\
   3 & better than most sequels for tv movies i hate & love (15) good (12) character (11) life (11)   & 0.174 \\
   & male because marie has a crush on her attractive & little (10) ever (9) watch (9) first (9) & \\\midrule
   & new <UNK> <UNK> via <UNK> <UNK> with absolutely hilarious & excellent (50) film (25) perfectly (19) wonderful (19) & \\
   4 & homosexual and an italian clown <UNK> is an entertaining & perfect (16) hilarious (15) best (13) fun (12)    & 0.141 \\
   & stephen <UNK> on the vampire <UNK> as a masterpiece & highly (11)  movie (11) brilliant (9)  old (9) & \\\midrule
 \bottomrule
\end{tabular}
}
\label{table:nlp}
\vspace{-3.5mm}
\end{table}
\paragraph{Setting:} We apply our method on IMDB, a text dataset with movie reviews classified as either positive or negative. 
We use 37500 reviews for training and 12500 for testing. We employ a 4-layer CNN model with 0.9 test accuracy. We apply our concept discover method to obtain 4 concepts, where the part of data $\x_j^i$ consists of 10 consecutive words of the sentence. The completeness of the 4 concepts is 0.97, thus the 4 concepts are highly representative of the classification model. 
\vspace{-3mm}
\paragraph{Result:} For each concept, Table \ref{table:nlp} shows (a) the top nearest neighbors based on the dot product of the concept and part of reviews (b) the most frequent words in the top-500 nearest neighbors (excluding stop words) (c) the conceptSHAP score for each concept. We can see that concepts 1 and 2 contain mostly negative sentiments, evident from the nearest neighbors -- concept 1 tends to criticize the movie/film directly, while concept 2 contains negativity in comments via words such as ``not'', ``doesn't'', ``even''. We note that the ratings in concept 2 are also negative since the scores 1 and 2 are considered to be very negative in movie review. On the other hand, concepts 3 and 4 contain mostly positive sentiments, as evident from the nearest neighbors -- concept 3 seems to discuss the plot of the movie without directing acclaiming or criticizing the movie, while concept 4 often contains very positive adjectives such as ``excellent'', ``wonderful'' that are extremely positive. More nearest neighbors are provided in the Appendix.

\vspace{-3mm}
\paragraph{Appending discovered concepts:} We perform an additional experiment where we randomly append 5 nearest neighbors (out of 500-nearest neighbors) of each concept to the end of all testing instances for further validation of the usefulness of the discovered concepts. For example, we may add ``wasting my time with a comment but this movie'' along with 4 other nearest neighbors of concept 1 to the end of a testing sentence. The original average prediction score for the testing sentences is 0.516, and the average prediction score after randomly appending 5 nearest neighbors of each concept becomes 0.103, 0.364, 0.594, 0.678 for concept 1, 2, 3, 4. As a controlled experiment, we appended 5 random sentences to the testing sentences, and the average prediction score is 0.498. This suggests that the concept score is highly related to the how the model makes prediction and may be used to manipulate the prediction. We note that while concept 1 contains stronger and more direct negative words than concept 2, concept 2 has a higher conceptSHAP value than concept 1. We hypothesize this is due to the fact that concept 2 may better detect weak negative sentences that may be difficult to be explained by concept 1, and thus may contribute more to the completeness score.

\vspace{-2mm}
\section{Conclusions}
\vspace{-2mm}

We propose to quantify the sufficiency of a particular set of concepts in explaining the model's behavior by the \emph{completeness} of concepts. By optimizing the completeness term coupled with additional constraints to ensure interpretability, we can discover concepts that are complete and interpretable. Through experiments on synthetic and real-world image and language data, we demonstrate that our method can recover ground truth concepts correctly, and provide conceptual insights of the model by examining the nearest neighbors. Although our work focuses on post-hoc explainability of pre-trained DNNs, joint training with our proposed objective function is possible to train inherently-interpretable DNNs. An interesting future direction is exploring the benefits of joint learning of the concepts along with the model, for better interpretability.

\newpage
\section{Broader Impact}

Bringing explainability can be crucial for AI deployments, for decision makers to build trust, for users to understand decisions, and for model developers to improve the quality. There are many use cases, from Finance, Healthcare, Employment/Recruiting, Retail, Environmental Sciences etc., that the explainability indeed constitutes the bottleneck to use deep neural networks (DNNs) despite their high performance. Thus, bringing explainability to DNNs can open many horizons for new AI deployments.

There are different forms of explainability, and our contributions are specifically for the very important `concept-based' explanations, towards a coherent and complete transparency to DNNs.
Our paper contributes to the quantification of the ``completeness'' of concept explanations, which can be useful to evaluate how sufficient existing and future concept-based explanations are, on the task of explaining a neural network. Validating the how sufficient the explanations are in explaining the model is a necessary sanity check, but often overlooked for concept explanations. As the field of explainable AI progresses rapidly, critiques and doubts on whether explanations are actually useful and accountable for models have also increased. Our objective metric bridges the gap between existing explanations and model accountability.

Most post-hoc concept-based explanations are applied only on image data. Our method is data type agnostic.
We demonstrate our canonical idea on both image and language data, which we believe can be applied to other data types as well. We believe our work lays a groundwork on general concept-based explanations, and hopefully will encourage future works on exploring concept explanations on all kinds of data types. Our concept discovery method explain a model using a small number of concepts, which can be explained by providing nearest neighbors in the training data to help users understand the concepts better. Such explanations provide a broader understanding of the model compared to the widely-used methods, such as saliency maps, and can be helpful for model developers and data scientists in understanding how the model works, improving the model based on insights, and ultimately learning from AI systems to build better AI systems.

\section{Ackowlegement} We acknowledge the support of DARPA via FA87501720152.






\bibliographystyle{unsrtnat} 
\bibliography{icml2020_conference1.bib}

\begin{thebibliography}{48}
\providecommand{\natexlab}[1]{#1}
\providecommand{\url}[1]{\texttt{#1}}
\expandafter\ifx\csname urlstyle\endcsname\relax
  \providecommand{\doi}[1]{doi: #1}\else
  \providecommand{\doi}{doi: \begingroup \urlstyle{rm}\Url}\fi

\bibitem[Ribeiro et~al.(2016)Ribeiro, Singh, and Guestrin]{ribeiro2016should}
Marco~Tulio Ribeiro, Sameer Singh, and Carlos Guestrin.
\newblock Why should i trust you?: Explaining the predictions of any
  classifier.
\newblock In \emph{KDD}. ACM, 2016.

\bibitem[Lundberg and Lee(2017)]{lundberg2017unified}
Scott~M Lundberg and Su-In Lee.
\newblock A unified approach to interpreting model predictions.
\newblock In \emph{NIPS}, 2017.

\bibitem[Armstrong et~al.(1983)Armstrong, Gleitman, and
  Gleitman]{ARMSTRONG1983263}
Sharon~Lee Armstrong, Lila~R. Gleitman, and Henry Gleitman.
\newblock What some concepts might not be.
\newblock \emph{Cognition}, 13\penalty0 (3):\penalty0 263 -- 308, 1983.

\bibitem[Tenenbaum(1999)]{tenenbaum1999bayesian}
Joshua~Brett Tenenbaum.
\newblock \emph{A Bayesian framework for concept learning}.
\newblock PhD thesis, Massachusetts Institute of Technology, 1999.

\bibitem[Kim et~al.(2018)Kim, Wattenberg, Gilmer, Cai, Wexler, Viegas,
  et~al.]{kim2018interpretability}
Been Kim, Martin Wattenberg, Justin Gilmer, Carrie Cai, James Wexler, Fernanda
  Viegas, et~al.
\newblock Interpretability beyond feature attribution: Quantitative testing
  with concept activation vectors (tcav).
\newblock In \emph{ICML}, 2018.

\bibitem[Zhou et~al.(2018)Zhou, Sun, Bau, and Torralba]{zhou2018interpretable}
Bolei Zhou, Yiyou Sun, David Bau, and Antonio Torralba.
\newblock Interpretable basis decomposition for visual explanation.
\newblock In \emph{ECCV}, 2018.

\bibitem[Ghorbani et~al.(2019)Ghorbani, Wexler, Zou, and
  Kim]{ghorbani2019towards}
Amirata Ghorbani, James Wexler, James Zou, and Been Kim.
\newblock Towards automatic concept-based explanations.
\newblock \emph{NeurIPS}, 2019.

\bibitem[Bouchacourt and Denoyer(2019)]{bouchacourt2019educe}
Diane Bouchacourt and Ludovic Denoyer.
\newblock Educe: Explaining model decisions through unsupervised concepts
  extraction.
\newblock \emph{arXiv preprint arXiv:1905.11852}, 2019.

\bibitem[Gilpin et~al.(2018)Gilpin, Bau, Yuan, Bajwa, Specter, and
  Kagal]{gilpin2018explaining}
Leilani~H Gilpin, David Bau, Ben~Z Yuan, Ayesha Bajwa, Michael Specter, and
  Lalana Kagal.
\newblock Explaining explanations: An overview of interpretability of machine
  learning.
\newblock In \emph{2018 IEEE 5th International Conference on data science and
  advanced analytics}. IEEE, 2018.

\bibitem[Yang et~al.(2019)Yang, Du, and Hu]{yang2019evaluating}
Fan Yang, Mengnan Du, and Xia Hu.
\newblock Evaluating explanation without ground truth in interpretable machine
  learning.
\newblock \emph{arXiv preprint arXiv:1907.06831}, 2019.

\bibitem[Blei et~al.(2003)Blei, Ng, and Jordan]{blei2003latent}
David~M Blei, Andrew~Y Ng, and Michael~I Jordan.
\newblock Latent dirichlet allocation.
\newblock \emph{JMLR}, 3\penalty0 (Jan):\penalty0 993--1022, 2003.

\bibitem[Shapley(1988)]{shapley_1988}
Lloyd~S. Shapley.
\newblock \emph{A value for n-person games}, page 31–40.
\newblock 1988.

\bibitem[Smilkov et~al.(2017)Smilkov, Thorat, Kim, Vi{\'e}gas, and
  Wattenberg]{smilkov2017smoothgrad}
Daniel Smilkov, Nikhil Thorat, Been Kim, Fernanda Vi{\'e}gas, and Martin
  Wattenberg.
\newblock Smoothgrad: removing noise by adding noise.
\newblock \emph{arXiv preprint arXiv:1706.03825}, 2017.

\bibitem[Chen et~al.(2018)Chen, Song, Wainwright, and Jordan]{l_c_shapley}
Jianbo Chen, Le~Song, Martin~J. Wainwright, and Michael~I. Jordan.
\newblock L-shapley and c-shapley: Efficient model interpretation for
  structured data.
\newblock \emph{arXiv:1808.02610}, 2018.

\bibitem[Koh and Liang(2017)]{koh2017understanding}
Pang~Wei Koh and Percy Liang.
\newblock Understanding black-box predictions via influence functions.
\newblock In \emph{ICML}, 2017.

\bibitem[Yeh et~al.(2018)Yeh, Kim, Yen, and Ravikumar]{yeh2018representer}
Chih-Kuan Yeh, Joon Kim, Ian En-Hsu Yen, and Pradeep~K Ravikumar.
\newblock Representer point selection for explaining deep neural networks.
\newblock In \emph{NIPS}, 2018.

\bibitem[Khanna et~al.(2019)Khanna, Kim, Ghosh, and
  Koyejo]{khanna2019interpreting}
Rajiv Khanna, Been Kim, Joydeep Ghosh, and Sanmi Koyejo.
\newblock Interpreting black box predictions using fisher kernels.
\newblock In \emph{The 22nd International Conference on Artificial Intelligence
  and Statistics}, pages 3382--3390, 2019.

\bibitem[Arik and Pfister(2019)]{attention_prototypical}
Sercan~{\"{O}}. Arik and Tomas Pfister.
\newblock Attention-based prototypical learning towards interpretable,
  confident and robust deep neural networks.
\newblock \emph{arXiv:1902.06292}, 2019.

\bibitem[Wachter et~al.(2017)Wachter, Mittelstadt, and
  Russell]{Wachter2017Counterfactual}
S.~Wachter, Brent~D. Mittelstadt, and Chris Russell.
\newblock Counterfactual explanations without opening the black box: Automated
  decisions and the gdpr.
\newblock \emph{European Economics: Microeconomics \& Industrial Organization
  eJournal}, 2017.

\bibitem[Dhurandhar et~al.(2018)Dhurandhar, Chen, Luss, Tu, Ting, Shanmugam,
  and Das]{dhurandhar2018explanations}
Amit Dhurandhar, Pin-Yu Chen, Ronny Luss, Chun-Chen Tu, Paishun Ting,
  Karthikeyan Shanmugam, and Payel Das.
\newblock Explanations based on the missing: Towards contrastive explanations
  with pertinent negatives.
\newblock In \emph{Advances in Neural Information Processing Systems}, pages
  592--603, 2018.

\bibitem[Hendricks et~al.(2018)Hendricks, Hu, Darrell, and
  Akata]{Hendricks2018GroundingVE}
Lisa~Anne Hendricks, Ronghang Hu, Trevor Darrell, and Zeynep Akata.
\newblock Grounding visual explanations.
\newblock In \emph{ECCV}, 2018.

\bibitem[van~der Waa et~al.(2018)van~der Waa, Robeer, van Diggelen, Brinkhuis,
  and Neerincx]{vanderwaa2018}
Jasper van~der Waa, Marcel Robeer, Jurriaan van Diggelen, Matthieu Brinkhuis,
  and Mark Neerincx.
\newblock {Contrastive Explanations with Local Foil Trees}.
\newblock In \emph{2018 Workshop on Human Interpretability in Machine Learning
  (WHI)}, 2018.

\bibitem[Goyal et~al.(2019{\natexlab{a}})Goyal, Wu, Ernst, Batra, Parikh, and
  Lee]{goyal2019counterfactual}
Yash Goyal, Ziyan Wu, Jan Ernst, Dhruv Batra, Devi Parikh, and Stefan Lee.
\newblock Counterfactual visual explanations.
\newblock In \emph{International Conference on Machine Learning}, pages
  2376--2384, 2019{\natexlab{a}}.

\bibitem[Joshi et~al.(2019)Joshi, Koyejo, Vijitbenjaronk, Kim, and
  Ghosh]{Joshi2019Towards}
S.~Joshi, O.~Koyejo, Warut~D. Vijitbenjaronk, Been Kim, and Joydeep Ghosh.
\newblock Towards realistic individual recourse and actionable explanations in
  black-box decision making systems.
\newblock \emph{ArXiv}, abs/1907.09615, 2019.

\bibitem[Poyiadzi et~al.(2020)Poyiadzi, Sokol, Santos-Rodr{\'i}guez, Bie, and
  Flach]{Poyiadzi2020FACE}
Rafael Poyiadzi, Kacper Sokol, Ra{\'u}l Santos-Rodr{\'i}guez, T.~D. Bie, and
  Peter~A. Flach.
\newblock Face: Feasible and actionable counterfactual explanations.
\newblock \emph{Proceedings of the AAAI/ACM Conference on AI, Ethics, and
  Society}, 2020.

\bibitem[Lage et~al.(2019)Lage, Chen, He, Narayanan, Kim, Gershman, and
  Doshi-Velez]{lage2019human}
Isaac Lage, Emily Chen, Jeffrey He, Menaka Narayanan, Been Kim, Samuel~J
  Gershman, and Finale Doshi-Velez.
\newblock Human evaluation of models built for interpretability.
\newblock In \emph{Proceedings of the AAAI Conference on Human Computation and
  Crowdsourcing}, volume~7, pages 59--67, 2019.

\bibitem[Samek et~al.(2016)Samek, Binder, Montavon, Lapuschkin, and
  M{\"u}ller]{samek2016evaluating}
Wojciech Samek, Alexander Binder, Gr{\'e}goire Montavon, Sebastian Lapuschkin,
  and Klaus-Robert M{\"u}ller.
\newblock Evaluating the visualization of what a deep neural network has
  learned.
\newblock \emph{IEEE transactions on neural networks and learning systems},
  28\penalty0 (11):\penalty0 2660--2673, 2016.

\bibitem[Kim et~al.(2016)Kim, Khanna, and Koyejo]{kim2016examples}
Been Kim, Rajiv Khanna, and Oluwasanmi~O Koyejo.
\newblock Examples are not enough, learn to criticize! criticism for
  interpretability.
\newblock In \emph{NIPS}, 2016.

\bibitem[Ancona et~al.(2017)Ancona, Ceolini, {\"O}ztireli, and
  Gross]{ancona2017towards}
Marco Ancona, Enea Ceolini, Cengiz {\"O}ztireli, and Markus Gross.
\newblock Towards better understanding of gradient-based attribution methods
  for deep neural networks.
\newblock \emph{arXiv preprint arXiv:1711.06104}, 2017.

\bibitem[Yeh et~al.(2019)Yeh, Hsieh, Suggala, Inouye, and Ravikumar]{yeh2019on}
Chih-Kuan Yeh, Cheng-Yu Hsieh, Arun~Sai Suggala, David~I. Inouye, and Pradeep
  Ravikumar.
\newblock On the (in)fidelity and sensitivity of explanations.
\newblock In \emph{NeurIPS}, 2019.

\bibitem[Guidotti et~al.(2018)Guidotti, Monreale, Ruggieri, Turini, Giannotti,
  and Pedreschi]{guidotti2018survey}
Riccardo Guidotti, Anna Monreale, Salvatore Ruggieri, Franco Turini, Fosca
  Giannotti, and Dino Pedreschi.
\newblock A survey of methods for explaining black box models.
\newblock \emph{ACM computing surveys (CSUR)}, 51\penalty0 (5):\penalty0 1--42,
  2018.

\bibitem[Hooker et~al.(2018)Hooker, Erhan, Kindermans, and
  Kim]{hooker2018evaluating}
Sara Hooker, Dumitru Erhan, Pieter-Jan Kindermans, and Been Kim.
\newblock Evaluating feature importance estimates.
\newblock \emph{arXiv preprint arXiv:1806.10758}, 2018.

\bibitem[Yang and Kim(2019)]{yang2019benchmarking}
Mengjiao Yang and Been Kim.
\newblock Benchmarking attribution methods with relative feature importance.
\newblock \emph{arXiv}, pages arXiv--1907, 2019.

\bibitem[Chan et~al.(2015)Chan, Jia, Gao, Lu, Zeng, and Ma]{chan2015pcanet}
Tsung-Han Chan, Kui Jia, Shenghua Gao, Jiwen Lu, Zinan Zeng, and Yi~Ma.
\newblock Pcanet: A simple deep learning baseline for image classification?
\newblock \emph{IEEE transactions on image processing}, 24\penalty0
  (12):\penalty0 5017--5032, 2015.

\bibitem[Kingma and Welling(2013)]{kingma2013auto}
Diederik~P Kingma and Max Welling.
\newblock Auto-encoding variational bayes.
\newblock \emph{arXiv preprint arXiv:1312.6114}, 2013.

\bibitem[Chorowski et~al.(2019)Chorowski, Weiss, Bengio, and
  Oord]{chorowski2019unsupervised}
Jan Chorowski, Ron~J Weiss, Samy Bengio, and A{\"a}ron van~den Oord.
\newblock Unsupervised speech representation learning using wavenet
  autoencoders.
\newblock \emph{arXiv preprint arXiv:1901.08810}, 2019.

\bibitem[Radford et~al.(2017)Radford, J{\'{o}}zefowicz, and
  Sutskever]{unsupervised_sentiment}
Alec Radford, Rafal J{\'{o}}zefowicz, and Ilya Sutskever.
\newblock Learning to generate reviews and discovering sentiment.
\newblock \emph{arXiv:1704.01444}, 2017.

\bibitem[{Grover} et~al.(2019){Grover}, {Pulice}, {Simari}, and
  {Subrahmanian}]{8668423}
S.~{Grover}, C.~{Pulice}, G.~I. {Simari}, and V.~S. {Subrahmanian}.
\newblock Beef: Balanced english explanations of forecasts.
\newblock \emph{IEEE Transactions on Computational Social Systems}, 6\penalty0
  (2):\penalty0 350--364, 2019.

\bibitem[Locatello et~al.(2018)Locatello, Bauer, Lucic, R{\"a}tsch, Gelly,
  Sch{\"o}lkopf, and Bachem]{locatello2018challenging}
Francesco Locatello, Stefan Bauer, Mario Lucic, Gunnar R{\"a}tsch, Sylvain
  Gelly, Bernhard Sch{\"o}lkopf, and Olivier Bachem.
\newblock Challenging common assumptions in the unsupervised learning of
  disentangled representations.
\newblock \emph{arXiv preprint arXiv:1811.12359}, 2018.

\bibitem[Chen et~al.(2019)Chen, Li, Tao, Barnett, Rudin, and Su]{chen2019looks}
Chaofan Chen, Oscar Li, Daniel Tao, Alina Barnett, Cynthia Rudin, and
  Jonathan~K Su.
\newblock This looks like that: deep learning for interpretable image
  recognition.
\newblock In \emph{Advances in Neural Information Processing Systems}, pages
  8928--8939, 2019.

\bibitem[Koh et~al.(2020)Koh, Nguyen, Tang, Mussmann, Pierson, Kim, and
  Liang]{koh2020concept}
Pang~Wei Koh, Thao Nguyen, Yew~Siang Tang, Stephen Mussmann, Emma Pierson, Been
  Kim, and Percy Liang.
\newblock Concept bottleneck models.
\newblock \emph{arXiv preprint arXiv:2007.04612}, 2020.

\bibitem[Goyal et~al.(2019{\natexlab{b}})Goyal, Feder, Shalit, and
  Kim]{goyal2019explaining}
Yash Goyal, Amir Feder, Uri Shalit, and Been Kim.
\newblock Explaining classifiers with causal concept effect (cace).
\newblock \emph{arXiv preprint arXiv:1907.07165}, 2019{\natexlab{b}}.

\bibitem[Araujo et~al.(2019)Araujo, Norris, and Sim]{araujo2019computing}
Andr{\'e} Araujo, Wade Norris, and Jack Sim.
\newblock Computing receptive fields of convolutional neural networks.
\newblock \emph{Distill}, 4\penalty0 (11):\penalty0 e21, 2019.

\bibitem[Mcauliffe and Blei(2008)]{mcauliffe2008supervised}
Jon~D Mcauliffe and David~M Blei.
\newblock Supervised topic models.
\newblock In \emph{NIPS}, 2008.

\bibitem[Lampert et~al.(2009)Lampert, Nickisch, and
  Harmeling]{lampert2009learning}
Christoph~H Lampert, Hannes Nickisch, and Stefan Harmeling.
\newblock Learning to detect unseen object classes by between-class attribute
  transfer.
\newblock In \emph{CVPR}. IEEE, 2009.

\bibitem[Szegedy et~al.(2016)Szegedy, Vanhoucke, Ioffe, Shlens, and
  Wojna]{szegedy2016rethinking}
Christian Szegedy, Vincent Vanhoucke, Sergey Ioffe, Jon Shlens, and Zbigniew
  Wojna.
\newblock Rethinking the inception architecture for computer vision.
\newblock In \emph{CVPR}, 2016.

\bibitem[{Narodytska} and {Kasiviswanathan}(2017)]{adversarial_attack}
N.~{Narodytska} and S.~{Kasiviswanathan}.
\newblock Simple black-box adversarial attacks on deep neural networks.
\newblock In \emph{CVPR Workshops}, 2017.

\bibitem[Fujimoto et~al.(2006)Fujimoto, Kojadinovic, and
  Marichal]{fujimoto2006axiomatic}
Katsushige Fujimoto, Ivan Kojadinovic, and Jean-Luc Marichal.
\newblock Axiomatic characterizations of probabilistic and
  cardinal-probabilistic interaction indices.
\newblock \emph{Games and Economic Behavior}, 55\penalty0 (1):\penalty0 72--99,
  2006.

\end{thebibliography}
\newpage
\onecolumn
\begin{appendices}
\section{Relation to PCA} \label{sec:PCA}

\paragraph{PCA:}
We show that under strict conditions, the PCA vectors applied on an intermediate layer where the principle components are used as concept vectors, maximizes the completeness score.

\begin{proposition} \label{pro:pca_comp1}
When $h$ is an isometry function that maps from $(\Phi(\cdot),\|\cdot\|_F) \rightarrow (f(\cdot),\|\cdot\|_F) $, and additionally $\p(\x_i) = \mathbbm{1}[y_i], \; \forall (\x_i,y_i) \in V$ (i.e. the loss is minimized, $\mathbbm{1}[y_i]$ is the one hot vector of class $y_i$), and also assume $T=1$, $\E[z] = \langle \f(\x) , \con\rangle$, and $l$ is a linear function, the first $m$ PCA vectors maximizes the L2 surrogate of $\Com$. 
\end{proposition}

We note that the assumptions for this proposition are extremely stringent, and may not hold in general. When the isometry and other assumptions do not hold, PCA no longer maximizes the completeness score as the lowest reconstruction in the intermediate layer do not imply the highest prediction accuracy in the output. In fact, DNNs are shown to be very sensitive to small perturbations in the input \citep{adversarial_attack} -- they can yield very different outputs although the difference in the input is small (and often perceptually hard to recognize to humans).
Thus, even though the reconstruction loss between two inputs are low at an intermediate layer, subsequent deep nonlinear processing may cause them to diverge significantly. The principal components are also not trained to be semantically meaningful, but to greedily minimize the reconstruction error (or maximize the projected variance). Even though completeness score and PCA share the idea of minimizing the reconstruction loss via dimensionality reduction, the lack of human interpretability of the principle components is a major bottleneck for PCA.

We provide this proposition only because completeness and PCA share the idea of minimizing the reconstruction loss via dimensionality reduction.
Another notable limitation of using PCA as concept vectors is the lack of human interpretability of the principle components. The PCA vectors are not trained to be semantically meaningful, but to greedily minimize the reconstruction error (or maximize the projected variance). 
\paragraph{Proof of Proposition \ref{pro:pca_comp1}}
\begin{proof}
By the basic properties of PCA, the first $m$ PCA vectors (principal components) minimize the reconstruction $\ell_2$ error. Define the concatenation of the $m$ PCA vectors as a matrix $\conp$ and $\| \cdot\|$ as the $\ell_2$ norm, and define $\text{proj}(\phi(\x, \conp))$ as the projection of $\x$ onto the span of $\conp$, the basic properties of PCA is equivalent to that for all $\con  = \begin{bmatrix} \con_1 \ \con_2 \hdots \ \con_m \end{bmatrix}$ , $$\sum_{\x \subseteq V_X}\|\text{proj}(\f(\x),\conp)-\f(\x)\|_F^2 \leq \sum_{\x \subseteq V_X} \|\text{proj}(\f(\x),\con)-\f(\x)\|_F^2.$$ 

By the isometry of $h$, we have $$ \sum_{\x \subseteq V_X}\|\h(\text{proj}(\f(\x),\conp))-\h(\f(\x))\|_F^2 \leq  \sum_{\x \subseteq V_X} \|\h(\text{proj}(\f(\x),\con))-\h(\f(\x))\|_F^2,$$ and since $\p(\x)$ is equal to Y, we can rewrite to
\begin{equation} \label{eq:last}
    \sum_{\x,y \subseteq V} \|\h(\text{proj}(\f(\x),\conp))-\mathbbm{1}[y]\|_F^2 \leq \sum_{\x,y \subseteq V} \|\h(\text{proj}(\f(\x),\con))-\mathbbm{1}[y]\|_F^2.
\end{equation}

We note that under the assumptions, $\E[\z|\x] = \f(\x) \con$, and thus the reconstruction layer $l$ can be written as 
\begin{equation}
    \begin{split}
        l & = \argmax_l \sum_{\x,y \subseteq V} \|\mathbbm{1}[y]- h(l(\E[\z|\x]))\|_F^2\\ & = \argmax_l \sum_{\x,y \subseteq V} \|\mathbbm{1}[y]- h(l(\f(\x) \con))\|_F^2\\ & = \argmax_l \sum_{\x \subseteq V_x} \|\f(\x)- l(\f(\x) \con)\|_F^2,
    \end{split}
\end{equation}

By definition, $\sum_{\x \subseteq V_x} \|\f(\x)- l(\f(\x) \con)\|_F^2$ is minimized by the projection, and thus $l(\f(\x) \con) = \text{proj}(\f(\x),\con)$.

And thus, \eqref{eq:last} can be written as:
\begin{equation*} 
    \sum_{\x,y \subseteq V} \|\h(l(\f(\x) \conp))-\mathbbm{1}[y]\|_F^2 \leq \sum_{\x,y \subseteq V} \|\h(l(\f(\x) \con))-\mathbbm{1}[y]\|_F^2.
\end{equation*}

and subsequently get that for any $\con$
$$ \frac{\mathbb{E}_{\x,y \sim V}[\|\mathbbm{1}[y] -P(y'|\E[z_{1:T}],h, \conp)\|_F^2]-R}{\mathbb{E}_{\x,y \sim V}[\|\mathbbm{1}[y] -P(\x_{1:T}, \p)\|_F^2]-R}  \geq  \frac{\mathbb{E}_{\x,y \sim V}[\|\mathbbm{1}[y] -P(y'|\E[z_{1:T}],h, \con)\|_F^2]-R}{\mathbb{E}_{\x,y \sim V}[\|\mathbbm{1}[y] -P(\x_{1:T}, \p)\|_F^2]-R}  .$$
\end{proof}

Thus, PCA vectors maximize the L2 surrogate of the completeness score. We emphasize that Proposition \ref{pro:pca_comp1} has several assumptions that may not be practical. However, the proposition is only meant to show that PCA optimizes our definition of completeness under a very stringent condition, as the key idea of completeness and PCA are both to prevent information loss through dimension reduction.

\section{Shapley Axioms for ConceptSHAP}

The axiomatic properties for ConceptSHAP are listed in the following proposition:
\begin{proposition}
    Given a set of concepts $C_S = \{\con_1, \con_2, ... \con_{m}\}$ and a completeness score $\Com$, and some importance score $\s_i$ for each concept $\con_i$ that depends on the completeness score $\Com$. $\s_i$ defined by conceptSHAP is the unique importance assignment that satisfy the following four axioms:
    \begin{itemize}
        \item Efficiency: The sum of all importance value should sum up to the total completeness score, $\sum_{i=1}^m \s_i(\Com) = \Com(C_S)$.
        \item Symmetry: For two concept that are equivalent s.t. $\Com(u\cup\{\con_i\}) = \Com(u\cup\{\con_j\}) $ for every subset $u\subseteq C_S \setminus \{\con_i,\con_j\}$, $\s_i(\Com) = \s_j(\Com)$.
        \item Dummy: If $\Com(u\cup\{\con_i\}) = \Com(u) $ for every subset $u\subseteq C_S \setminus \{\con_i\}$, then $\s_i(\Com) = 0 $.
        \item Additivity: If $\Com$ and $\Com'$ have importance value $\s(\Com)$ and $\s(\Com')$ respectively, then the importance value of the sum of two completeness score should be equal to the sum of the two importance values, i.e, $\s_i(\Com + \Com') = \s_i(\Com) + \s_i(\Com')$ for all i.
    \end{itemize}
\end{proposition}
The proof and the interpretation for these concepts are well discussed in \citep{shapley_1988, lundberg2017unified, fujimoto2006axiomatic}. 

\section{Additional Experiments Results and Settings}

\paragraph{Automated Alignment score on Synthetic Dataset}
Given the existence of each ground truth shape $\z_{1:5}^i$ in each sample $\x^i$, we can evaluate how closely the discovered concept vectors $\con_{1:m}$ align with the actual ground truth shapes 1 to 5. Our evaluation assumes that if $\con_k$ corresponds to some shape $v$, then the parts of input that contain the shape $v$ and the parts of input that does not contain ground truth shape $v$ can be linearly separated by $\con_k$. That is, $\con_k \cdot \x_a > \ \con_k \cdot \x_b$ or $\con_k \cdot \x_a < \ \con_k \cdot \x_b$ for all $\x_a$ that contains shape $v$ and all $\x_b$ that does not contain shape $v$. 
Without loss of generality, we assume $\con_k \cdot \x_a > \con_k \cdot \x_b$ if $\x_a$  contains shape $v$ and $\x_c^d$ does not contain shape $v$ 
for notation simplicity, and check $\con_k$ and $-\con_k$ for each discovered concepts. 
Following this assumption, $\max_{t =1}^T \con_k \cdot \x_t^i > \max_{t =1}^T \con_k \cdot \x_t^j$ for all $i,j$ such that $\z_{v}^i =1$ and $\z_{v}^j =0,$ since at least one part of $\x_t^b$ should contain the ground truth shape $v$.
Therefore, to evaluate how well $\con_k$ corresponds to shape $v$, we measure the accuracy of using $\mathbbm{1} [\max_{t =1}^T \con_k \cdot \x_t^i > \text{const}]$ to classify $\z_{v}^i$. More formally, we define the matching score between concept $\con_k$ to the shape v as:
\begin{equation*}
\begin{split}
\text{Match}(\con_k,\mathbf{\z}_v) =  \mathbb{E}_{\x^i \sim V}[\mathds{1}{[ \max_{t\subseteq [1,T]} \con_k \cdot \x_t^i > e] = \mathbf{\z}_{v}^i}],
\end{split}
\end{equation*}
where $e$ is some constant. We then evaluate how well the set of discovered concepts $\con_{1:m}$ aligns with shapes 1 to 5:
\begin{equation*}
\begin{split}
        \text{Alignment}(\con_{1:m},\z_{1:5}) = \! \max_{ P \in [1,m]^m}\frac{1}{5}\sum_{j=1}^{5} \text{Match}(\con_{P[j]},\mathbf{\z}_j),
\end{split}
\end{equation*}
which measures the best average matching accuracy by assigning the best concept vector to differentiate each shape. For each concept vector $\con_j$, we test $\con_j$ and $-\con_j$ and choose the direction that leads to the highest alignment score.
\paragraph{Creation of the Toy Example}

The complete list of the target y is $y_1 = \sim(\mathbf{\z_1} \cdot \mathbf{\z_3})+ \mathbf{\z_4}, y_2 =  \mathbf{\z_2}+ \mathbf{\z_3}+ \mathbf{\z_4}, y_3 =  \mathbf{\z_2}\cdot \mathbf{\z_3} + \mathbf{\z_4}\cdot \mathbf{\z_5}, y_4 = \mathbf{\z_2} \text{ XOR } \mathbf{\z_3}, y_5 = \mathbf{\z_2} + \mathbf{\z_5}, y_6 = \sim (\mathbf{\z_1} + \mathbf{\z_4}) + \mathbf{\z_5}, y_7 = (\mathbf{\z_2} \cdot \mathbf{\z_3}) \text{ XOR } \mathbf{\z_5}, y_8 = \mathbf{\z_1} \cdot \mathbf{\z_5}+\mathbf{\z_2}, y_9 =  \mathbf{\z_3} , y_{10} = (\mathbf{\z_1} \cdot \mathbf{\z_2}) \text{ XOR } \mathbf{\z_4}, y_{11} =  \sim(\mathbf{\z_3}+ \mathbf{\z_5}), y_{12} =  \mathbf{\z_1}+ \mathbf{\z_4}+ \mathbf{\z_5},  y_{13} =  \mathbf{\z_2} \text{ XOR } \mathbf{\z_3}, y_{14} = \sim(\mathbf{\z_1} \cdot \mathbf{\z_5} + \mathbf{\z_4}), y_{15} = \mathbf{\z_4}  \text{ XOR } \mathbf{\z_5}.$

We create the dataset in matplotlib, where the color of each shape is sampled independently from green, red, blue, black, orange, purple, yellow, and the location is sampled randomly with the constraint that different shapes do not coincide with each other.

\paragraph{Hyper-parameter Choice and Sensitivity}
To choose the hyperparameters, one can use a small-scale evaluation dataset to choose a few important hyperparameters. One should choose the hyperparamters so that they get concepts with high completeness and $R_1 (\mathbf{c})$, and we better describe the impact of these hyperparamters to guide such selection. 

\begin{figure} 
\centering
\vspace{-1mm}
  \includegraphics[width=0.5\linewidth]{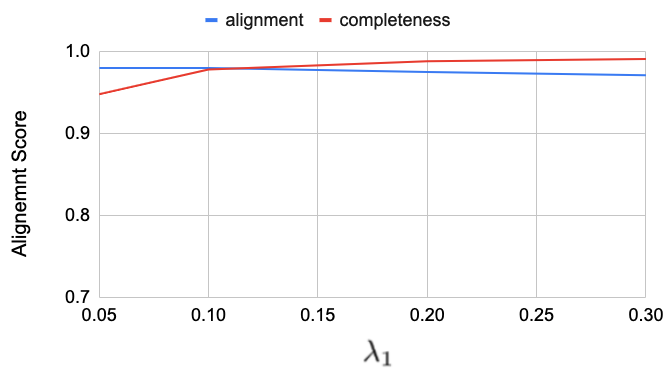}
  \caption{Completeness score and Alignment score for different hyper-parameter $\lambda_1$. }
  \vspace{-1mm}
 \label{fig:toy_para_1}
\end{figure}

\begin{figure} 
\centering
\vspace{-1mm}
  \includegraphics[width=0.5\linewidth]{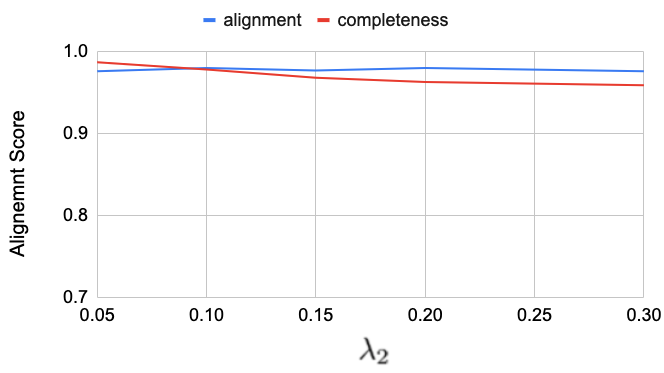}
  \caption{Completeness score and Alignment score for different hyper-parameter $\lambda_2$. }
  \vspace{-1mm}
 \label{fig:toy_para_2}
\end{figure}

\begin{figure} 
\centering
\vspace{-1mm}
  \includegraphics[width=0.5\linewidth]{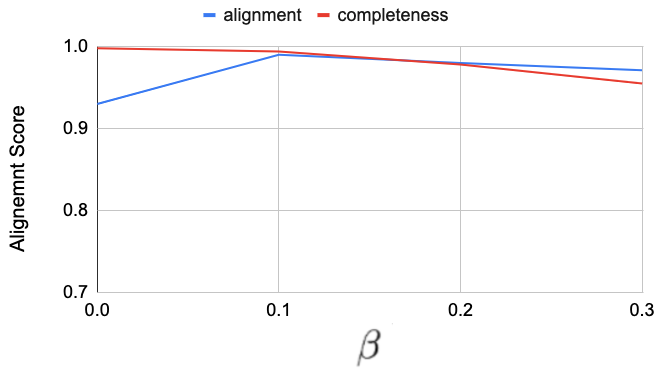}
  \caption{Completeness score and Alignment score for different hyper-parameter $\beta$. }
  \vspace{-1mm}
 \label{fig:toy_para_3}
\end{figure}

\textbf{Choice of $\mathbf{\lambda_1,\lambda_2, \beta}$:}
We set $ \lambda_1 = \lambda_2 = 0.1, \beta = 0.2$ for the toy dataset. We show the completeness score for varying $\lambda_1, \lambda_2,\beta$ in Figure \ref{fig:toy_para_1},\ref{fig:toy_para_2},\ref{fig:toy_para_3} (when varying $\lambda_1$, we fix $\lambda_2 = 0.1$, and $\beta = 0.2$.) We see that both the completeness and alignment score are above 0.93 when $ \lambda_1$ and $ \lambda_2$ are in the range of $[0.05,0.3]$, and $\beta$ is in the range of $[0,0.3]$, and thus our method outperforms all baselines with a wide range of hyper-parameters.  Therefore, our method is not sensitive to the hyper-parameter in the toy dataset. We set the $ \lambda_1 = \lambda_2 = 0.1, \beta = 0.3$ for the NLP dataset, and we set $ \lambda_1 = \lambda_2 = 10.0, \beta = 0$ for AwA dataset since the optimization becomes more difficult with a deeper neural network, and thus we increase the regularizer strength to ensure interpretability. The completeness is above 0.9 when $ \lambda_1$ and $ \lambda_2$ are set in the range of $[2,20]$. Overall, our method is not too sensitive to the selection of hyper-parameter. The general principle for hyper-parameter tuning is to chose larger $ \lambda_1$ and $ \lambda_2$ that still gives a completeness value (usually > 0.95).

\begin{figure}
  \centering
  \includegraphics[width=0.8\linewidth]{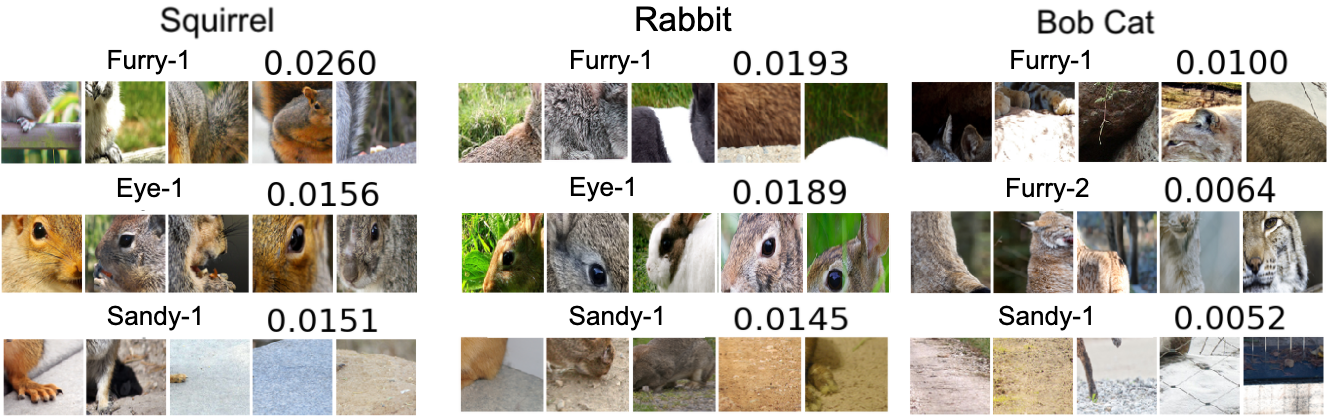}
  \vspace{-1mm}
  \caption{Nearest neighbors when applied to mixed\_5c layer. The discovered concepts focus on smaller patches and captures lower level.}
    \label{fig:nn_awa_smaller}
\end{figure}

\textbf{Choice of $h, g$:} The intermediate layer (which effects $h$) can be chosen depending on the size of the nearest neighbor patch the user would like to visualize, since this size depends on the receptive field of the feature layer. Deeper layers have larger receptive fields while shallower layers have smaller receptive field size. In AwA experiments, we choose Mixed\_5d layer so that the receptive field is $127 \times 127$, which is half of the original image size and capable of capturing both larger and smaller concepts. If the user is only interested in smaller and more low level concepts (such as the texture of image), we can apply our method to earlier layers. As a hyper-parameter control experiment, we apply our method to mixed\_5c layer where the receptive field is $95 \times 95$ and visualize the result in Fig. \ref{fig:nn_awa_smaller}, where indeed smaller (low level) concepts such as eyes (in constrast to head), furry, sandy are captured (which we name the concepts). If one finds the concepts discovered to be too small (low level), they could apply the method on a deeper layer and vice versa. For the choice of $g$, we let it be a two layer neural network (with 512 neurons) followed by the remaining network $h$, so that $h$ is also optimized in eq.3, which we fix in all experiments.

\begin{figure} 
\centering
\vspace{-1mm}
  \includegraphics[width=0.99\linewidth]{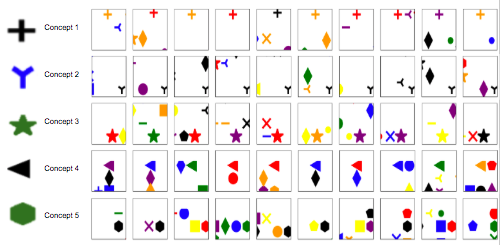}
  \caption{(Larger Version) Nearest Neighbors for each concept obtained in the toy example. }
  \vspace{-1mm}
 \label{fig:toy_large}
\end{figure}

\paragraph{Additional Nearest Neighbors for toy example}
We show 10 nearest neighbors for each concept obtained by our methods and baseline methods in the toy example in Figure \ref{fig:toy_large}. The 10 nearest neighbors for each concept obtained by different methods is used to perform the user study, to test if the nearest neighbors allow human to retrieve the correct ground truth concepts for each method.

\begin{figure} 
\centering
\vspace{-1mm}
  \includegraphics[width=0.7\linewidth]{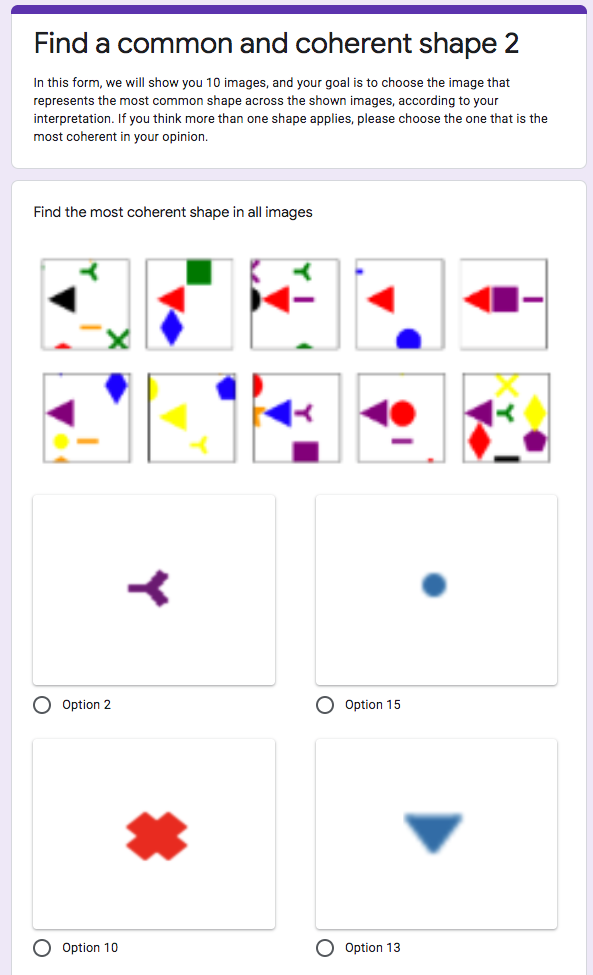}
  \caption{An example question of a screenshot of the human study. }
  \vspace{-1mm}
 \label{fig:user}
\end{figure}

\paragraph{User Study Setting and Discussion}
For the user study, we set $m=5$ (i.e. 5 discovered concepts) for all compared methods. The order of the 2 randomly chosen conditions (and which 2 conditions are paired), the order of the questions, the order of choices are all randomized to avoid biases and learning effects. All users are graduate students with some knowledge of machine learning. None of them have (self-reported) color-blindness. For each discovered concept, an user is asked to find the most common and coherent shape given the top 10 nearest neighbors. An example question is shown in Figure \ref{fig:user}. Each user is given 10 questions, which correspond to the nearest neighbors of the discovered concepts for two random methods. (each method has 5 discovered concepts, and thus two methods have 10 discovered concepts in total). There are 20 users in total, and thus each method is tested on 8 users. For each method, we report the average number of correct answers chosen by the users. For example, if an user chooses shape 1,2,5,7,5, then the number of the correct answers chosen by the user will be 3 (since 1,2,5 are the ground truth shape obtained by the user). We average the correct answers chosen by 8 users for each method to obtain the ``average number of correct answers chosen by users''. We also report the average number of agreed answers chosen by the users. For example, if most users choose 1,2,5,7,5 for five questions respectively, we set 1,2,5,7,5 as the ground truth for the five questions. If user A answered 1,2,5,7,10 for the five questions respectively, his number of agreed answers would be 4. We average the agreed answers chosen by 8 users for each method to obtain the ``average number of agreed answers chosen by users''.

We find that other methods mainly fail due to (a) the same concept are chosen repeatedly (e.g. concept 2 and concept 4 of ACE). (b) lack of disentanglement (coherency) of concepts (e.g. concept 5 of PCA shows two shape in all 10 nearest neighbors). (c) highlighted concepts are not related to the ground truth concept (e.g. concept 4 of Kmeans). (a), (c) are related to the lack of completeness of the method, and (b) is related to the lack of coherency of the method.

\paragraph{Implementation Details }
For calculating ConceptSHAP, we use the method in kernelSHAP \citep{lundberg2017unified} to calculate the Shapley values efficiently by regression. For ACE in toy example, we set the number of cluster to be 15, and choose the concepts based on TCAV score. For ACE in toy example, we set the number of clusters to be 150, and choose the concepts based on TCAV score. For PCA, we return the top $m$ principle components when the number of discovered concepts is $m$. For k-means, we set the cluster size to be $m$ when the number of discovered concepts is $m$, and return the cluster mean as the discovered concepts.

\begin{figure}[ht]
  \centering
  \includegraphics[width=0.8\linewidth]{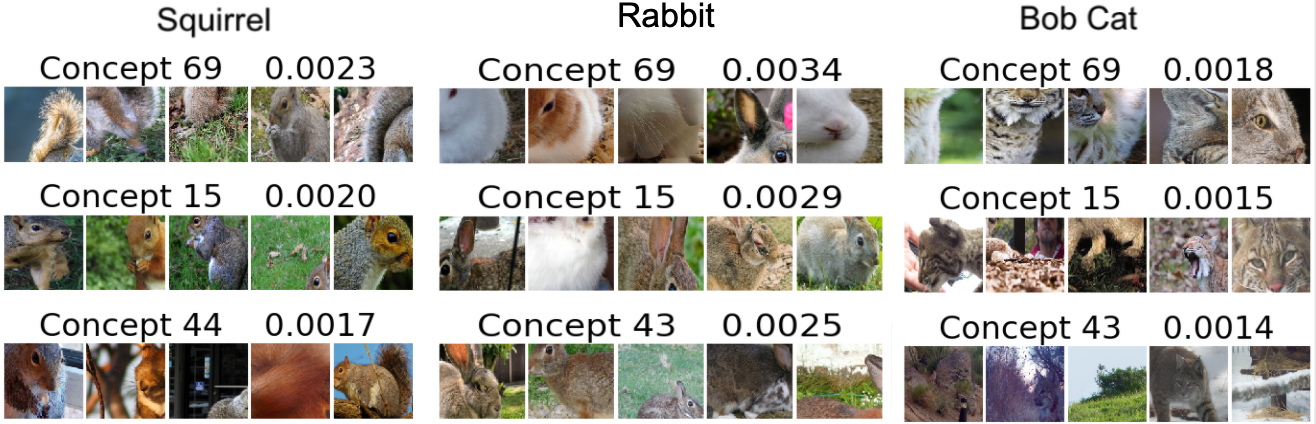}
  \vspace{-1mm}
  \caption{Nearest neighbors of top concepts by PCA.}
    \label{fig:nn_awa_pca}
\end{figure}

\begin{figure}[ht]
  \centering
  \includegraphics[width=0.8\linewidth]{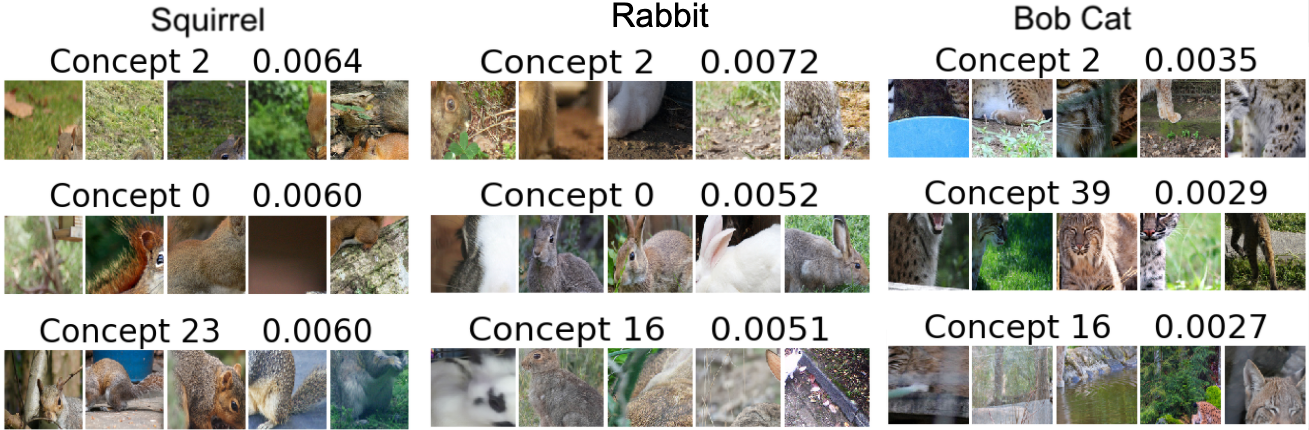}
  \vspace{-1mm}
  \caption{Nearest neighbors of top concepts by Kmeans.}
    \label{fig:nn_awa_kmeans}
\end{figure}

\paragraph{Qualitative Results for Baselines in AwA}
We show nearest neighbors of the top concepts of PCA and Kmeans of the three class ``Rabbit'', ``Squirrel'', and ``Weasel'' in AwA in figure \ref{fig:nn_awa_pca} and \ref{fig:nn_awa_kmeans} respectively.

\begin{figure} 
\centering
\vspace{-1mm}
  \includegraphics[width=1.1\linewidth]{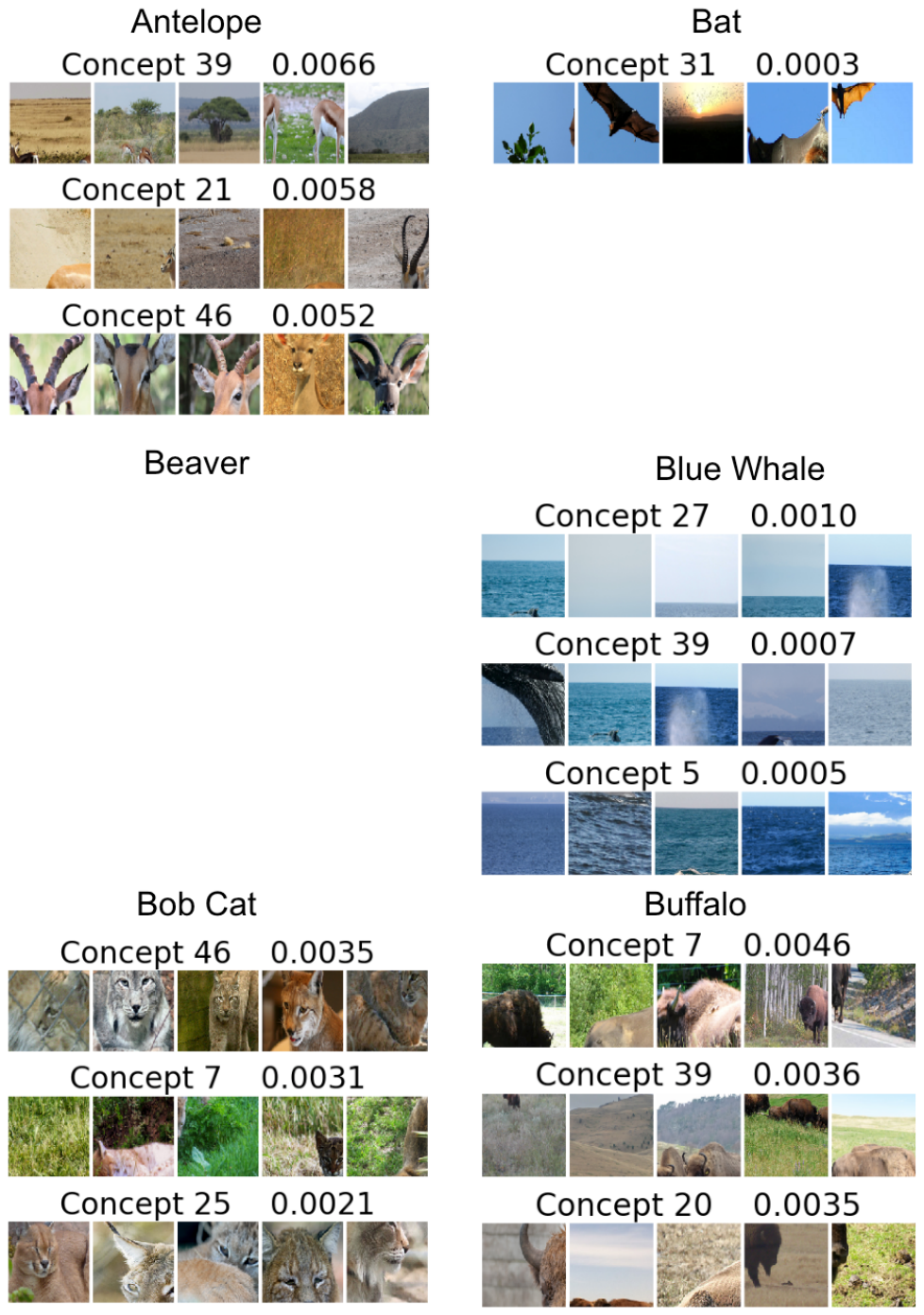}
 \vspace{-10mm}
  \caption{(Larger Version) Nearest Neighbors for each concept obtained in AwA. }

 \label{fig:awa_full_0}
\end{figure}

\begin{figure} 
\centering
\vspace{-1mm}
  \includegraphics[width=1.1\linewidth]{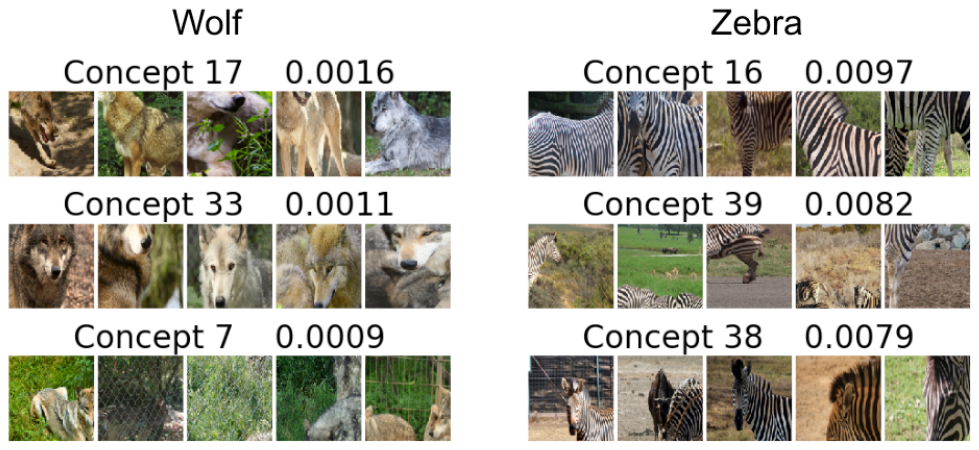}
   \vspace{-10mm}
  \caption{(Larger Version) Nearest Neighbors for each concept obtained in AwA. }
 \label{fig:awa_full_8}
\end{figure}

\paragraph{Additional Nearest Neighbors for AwA}
We show additional nearest neighbors of the top concepts in AwA for all 50 classes from Figure \ref{fig:awa_full_0} to Figure \ref{fig:awa_full_8}. For each class, the 3 concepts with the highest ConceptSHAP respect to the class with $R_1(\con)$ above 0.8 is shown, along with the ConceptSHAP score with respect to the class. We see that many important concepts are shared between different classes, where most of them are semantically meaningful. To list some examples, concept 7 corresponds to the concept of grass, concept 33 shows a specific kind of wolf-like face (which has two different colors on the face), concept 27 corresponds to the sky/ocean view, concept 25 shows a side face that is shared among many animals, concept 46 shows a front face of cat-like animals, concept 21 shows sandy/ wilderness texture of the background, concept 38 shows gray back ground that looks like asphalt road, concept 43 shows similar ears of several animals, concept 31 shows furry/ rough texture with a plain background. 

\paragraph{Additional Nearest Neighbors for NLP} We show additional nearest neighbors of the 4 concepts in NLP. The nearest neighbors of concept 1 and concept 2 are generally negative, and concept 3 and concept 4 are generally positive.

\clearpage

\begin{figure} 
\centering
\vspace{-1mm}
  \includegraphics[width=0.99\linewidth]{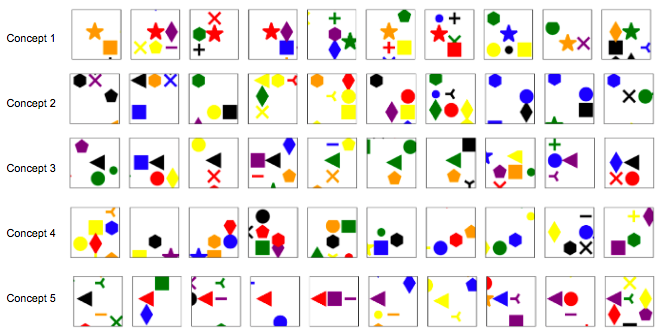}
  \caption{(Larger Version) Nearest Neighbors for each concept for ACE obtained in the toy example. }
  \vspace{-1mm}
 \label{fig:toy_large2}
\end{figure}
\begin{figure} 
\centering
\vspace{-1mm}
  \includegraphics[width=0.99\linewidth]{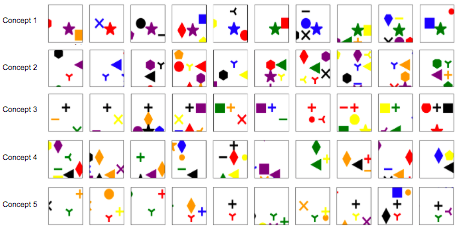}
  \caption{(Larger Version) Nearest Neighbors for each concept for PCA obtained in the toy example. }
  \vspace{-1mm}
 \label{fig:toy_large3}
\end{figure}
\begin{figure} 
\centering
\vspace{-1mm}
  \includegraphics[width=0.99\linewidth]{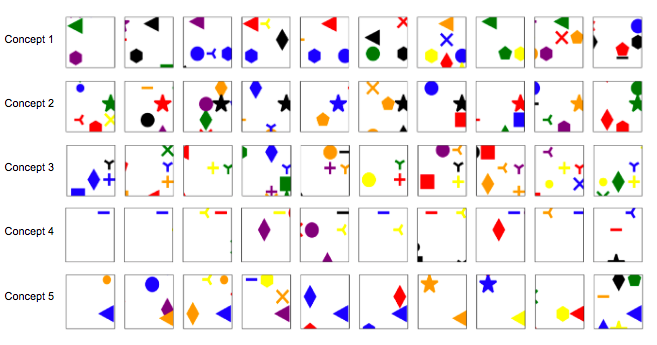}
  \caption{(Larger Version) Nearest Neighbors for each concept for Kmeans obtained in the toy example. }
  \vspace{-1mm}
 \label{fig:toy_large4}
\end{figure}

\begin{figure} 
\centering
\vspace{-1mm}
  \includegraphics[width=0.99\linewidth]{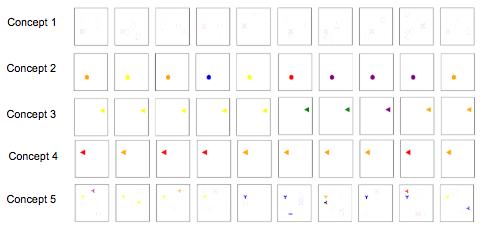}
  \caption{(Larger Version) Nearest Neighbors for each concept for ACE-SP obtained in the toy example. }
  \vspace{-1mm}
 \label{fig:toy_large5}
\end{figure}

\begin{table}[!t]
\caption{The 4 discovered concepts with more nearest neighbors.}
\centering
\adjustbox{max width=\linewidth}{
\centering
\small
\begin{tabular}{@{}lcc@{}}
\toprule
  Concept & Nearest Neighbors  \\ \midrule
   & poorly constructed what comes across as interesting is the  \\
    & wasting my time with a comment but this movie  \\
   & awful in my opinion there were <UNK> and the  \\
  1 & forgettable <UNK> earn far more critical acclaim and win  \\
   & wasting my time with a comment but this movie  \\
   & worst 80's slashers alongside <UNK> with fear <UNK> deadly  \\
   & worst ever sound effects ever used in a movie  \\
   \midrule
    & normally it would earn at least 2 or 3 \\
   & <UNK> <UNK> is just too dumb to be called  \\
   & i feel like i was ripped off and hollywood  \\
   2& johnson seems to be the only real actor here  \\
   & but this thing is watchable if only for bela  \\
   & performance but they're all too unlikable to really care  \\
   & way the fights are awfully bad done while sometimes  \\
   \midrule
   & remember awaiting return of the jedi with almost <UNK> \\
   & better than most sequels for tv movies i hate   \\
   & male because marie has a crush on her attractive \\
   3 & think that about a lot of movies in this   \\
   & i am beginning to see what she has been \\
   & cinema of today these films are the products of  \\
   & long last think eastern promises there will be blood \\
   \midrule
   & new <UNK> <UNK> via <UNK> <UNK> with absolutely hilarious  \\
    & homosexual and an italian clown <UNK> is an entertaining \\
   & stephen <UNK> on the vampire <UNK> as a masterpiece  \\
   4& and between the scenes the movie has one gem \\
   & make a film <UNK> in color light so perfectly  \\
   & in it and the evil beast is an incredible \\
   & father and my adult son peter falk is excellent   \\\midrule
 \bottomrule
\end{tabular}
}
\label{table:nlp_long}
\end{table}

\begin{figure}
\centering
\vspace{-1mm}
  \includegraphics[width=1.1\linewidth]{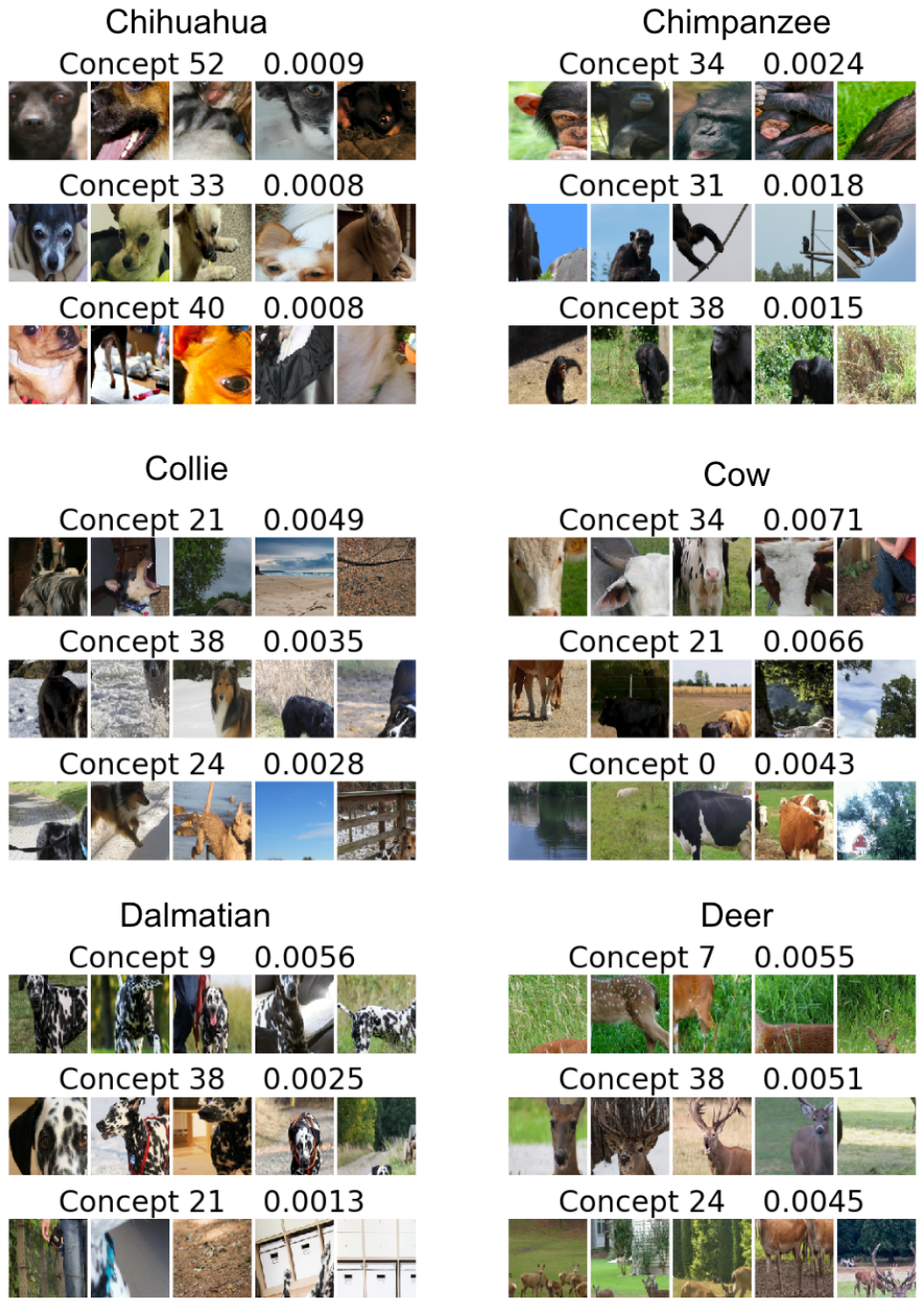}
   \vspace{-10mm}
  \caption{(Larger Version) Nearest Neighbors for each concept obtained in AwA. }

\end{figure}
\begin{figure}
\centering
\vspace{-1mm}
  \includegraphics[width=1.1\linewidth]{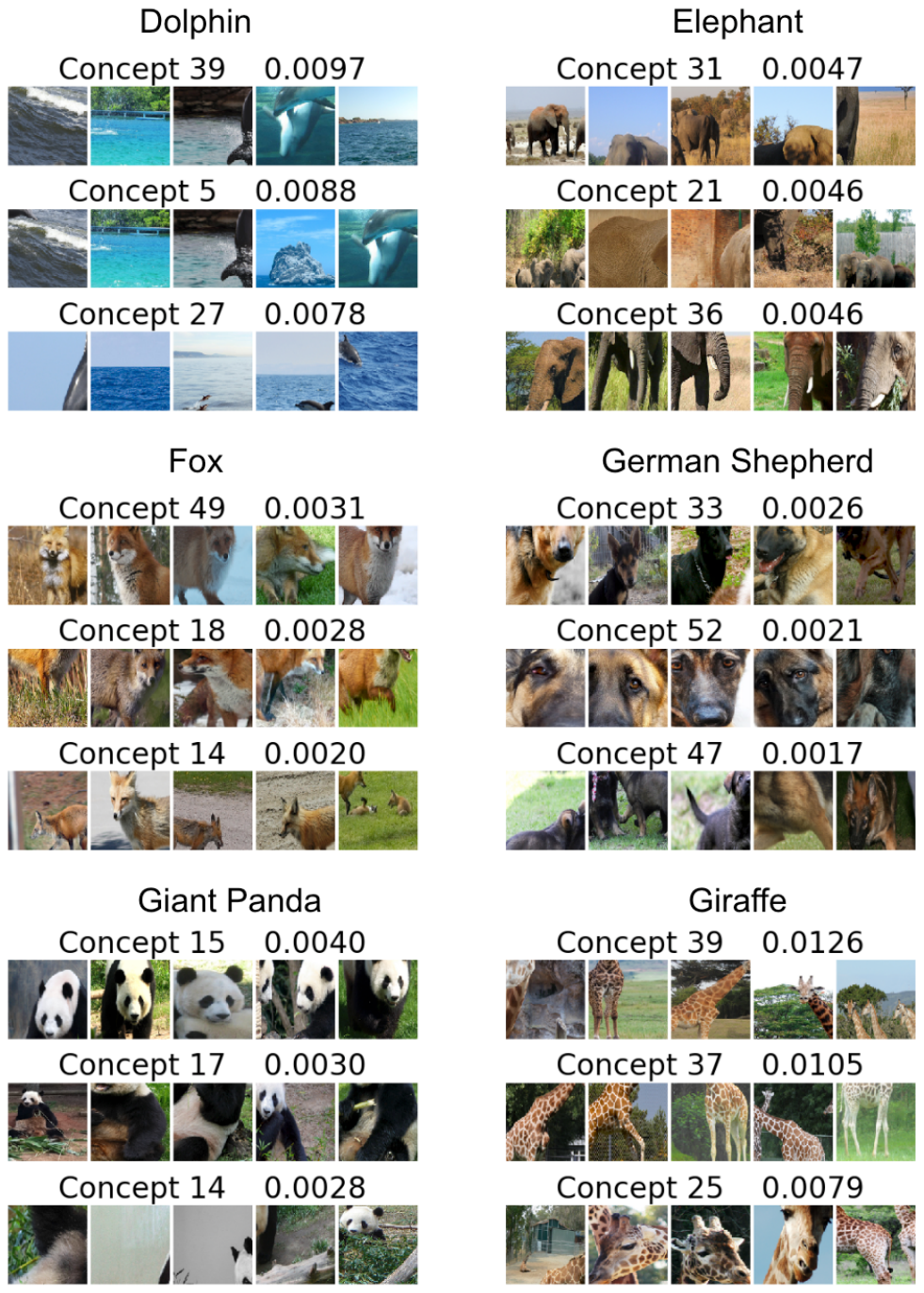}
   \vspace{-10mm}
  \caption{(Larger Version) Nearest Neighbors for each concept obtained in AwA. }

\end{figure}
\begin{figure}
\centering
\vspace{-1mm}
  \includegraphics[width=1.1\linewidth]{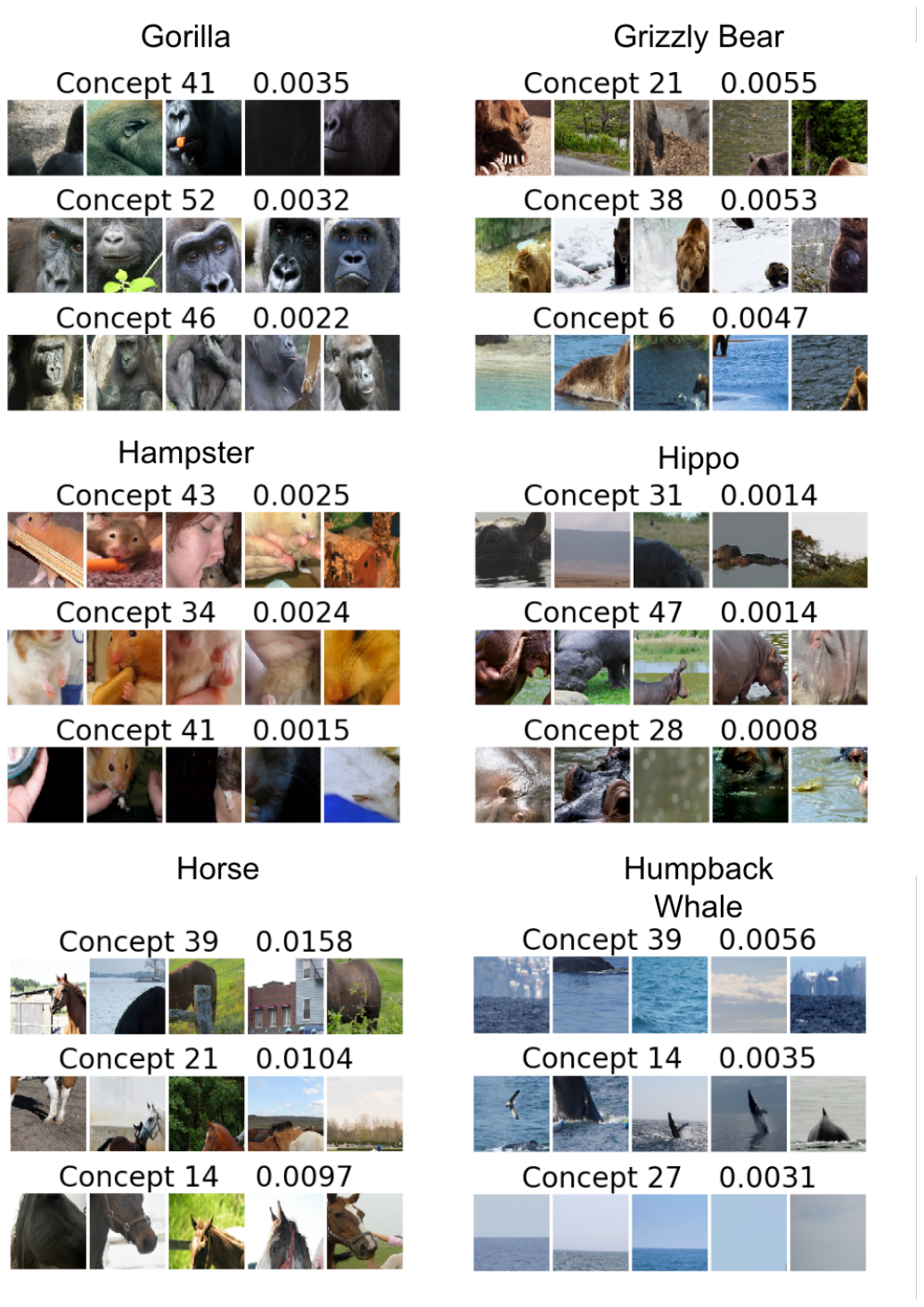}
   \vspace{-10mm}
  \caption{(Larger Version) Nearest Neighbors for each concept obtained in AwA. }
 
\end{figure}
\begin{figure}
\centering
\vspace{-1mm}
  \includegraphics[width=1.1\linewidth]{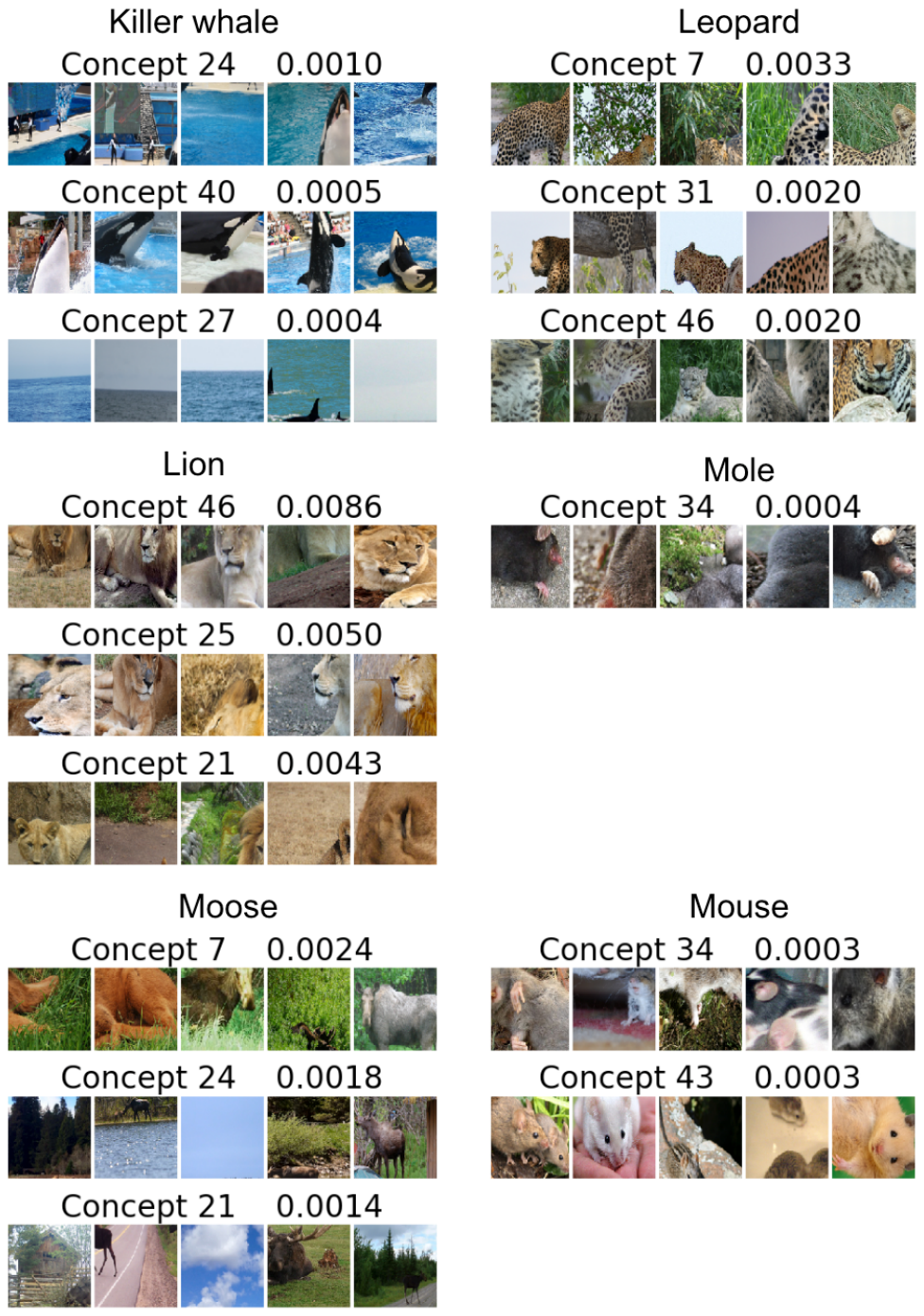}
   \vspace{-10mm}
  \caption{(Larger Version) Nearest Neighbors for each concept obtained in AwA. }
 
\end{figure}
\begin{figure}
\centering
\vspace{-1mm}
  \includegraphics[width=0.9\linewidth]{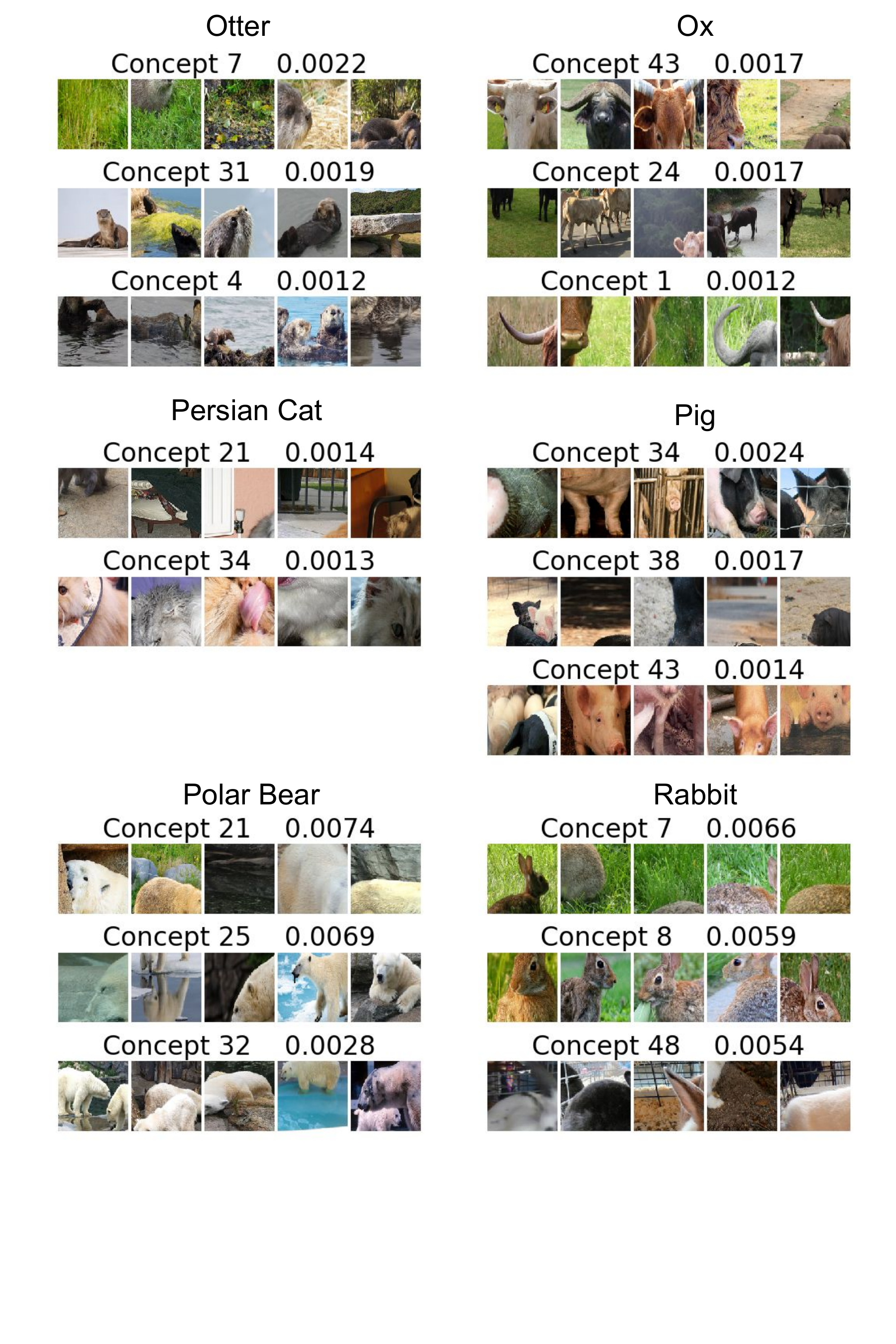}
   \vspace{-10mm}
  \caption{(Larger Version) Nearest Neighbors for each concept obtained in AwA. }
 
\end{figure}
\begin{figure}
\centering
\vspace{-1mm}
  \includegraphics[width=1.1\linewidth]{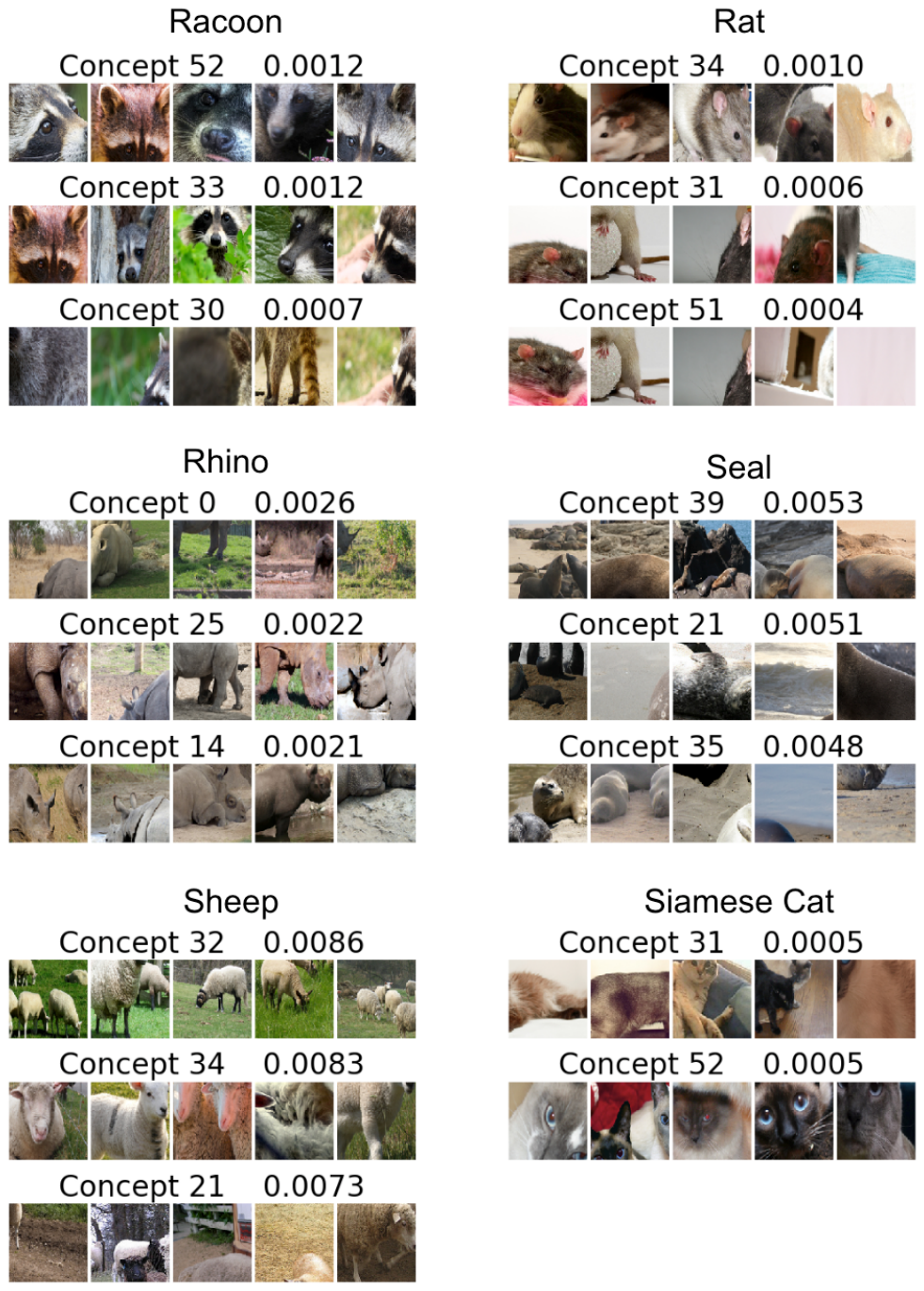}
   \vspace{-10mm}
  \caption{(Larger Version) Nearest Neighbors for each concept obtained in AwA. }
 
\end{figure}
\begin{figure}
\centering
\vspace{-1mm}
  \includegraphics[width=1.1\linewidth]{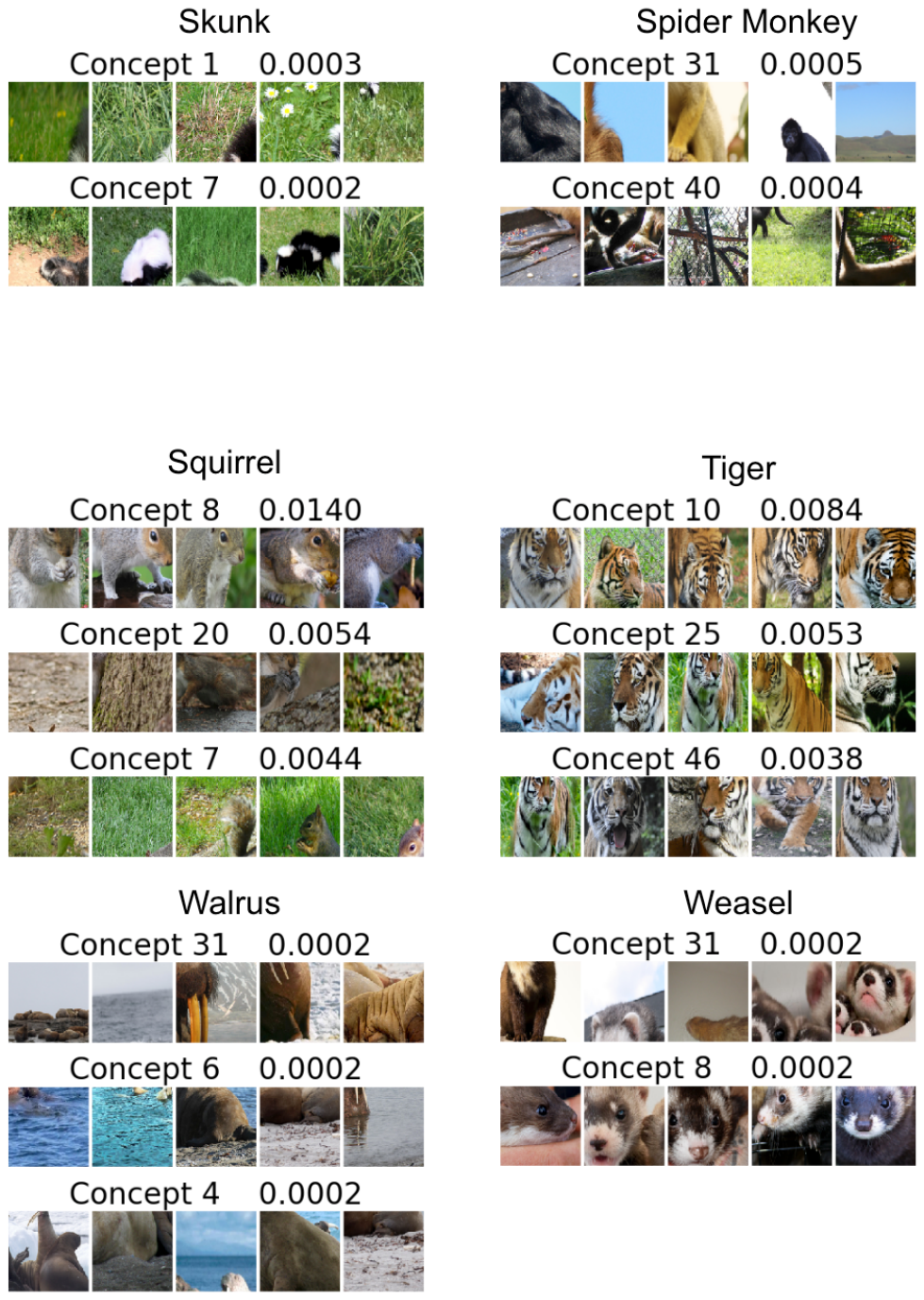}
   \vspace{-10mm}
  \caption{(Larger Version) Nearest Neighbors for each concept obtained in AwA. }
 
\end{figure}

\end{appendices}
\end{document}